\pdfoutput=1
\documentclass[runningheads]{llncs}
\usepackage{eccv}
\usepackage{eccvabbrv}
\usepackage{graphicx}
\usepackage{booktabs}
\usepackage[accsupp]{axessibility}
\usepackage{hyperref}
\usepackage{orcidlink}
\usepackage{algorithm}
\usepackage{algorithmic}
\usepackage{multicol}
\usepackage{multirow}
\usepackage{amsmath}

\begin{document}

\title{FMM-Attack: A Flow-based Multi-modal Adversarial Attack on Video-based LLMs} 
\titlerunning{FMM-Attack on Video-based LLMs}
\author{Jinmin Li\inst{1} \textsuperscript{$\star$} \and Kuofeng Gao\inst{1} \textsuperscript{$\star$} \and Yang Bai\inst{2} \textsuperscript{$\dagger$} \and \\ Jingyun Zhang \inst{3} \and Shu-tao Xia\inst{1,4} \and Yisen Wang \inst{5}}

\authorrunning{Li et al.}

\institute{
Tsinghua Shenzhen International Graduate School, Tsinghua University 
\and
Tencent Technology (Beijing) Co.Ltd 
\and
Tencent Lab 33
\and
Peng Cheng Laboratory
\and
Peking University
\\ \email{\{ljm22, gkf21\}@mails.tsinghua.edu.cn,\\
\{baiyang0522, zhang304973926\}@gmail.com \\
xiast@sz.tsinghua.edu.cn, yisen.wang@pku.edu.cn}
}
\maketitle

\renewcommand{\thefootnote}{\fnsymbol{footnote}} 
\footnotetext[1]{Equal contribution.}
\footnotetext[4]{Corresponding author. } 

\begin{abstract}
Despite the remarkable performance of video-based large language models (LLMs), their adversarial threat remains unexplored. To fill this gap, we propose the first adversarial attack tailored for video-based LLMs by crafting flow-based multi-modal adversarial perturbations on a small fraction of frames within a video, dubbed \textbf{FMM-Attack}. Extensive experiments show that our attack can effectively induce video-based LLMs to generate \textit{incorrect} answers when videos are added with imperceptible adversarial perturbations. Intriguingly, our FMM-Attack can also induce garbling in the model output, prompting video-based LLMs to hallucinate. Overall, our observations inspire a further understanding of multi-modal robustness and safety-related feature alignment across different modalities, which is of great importance for various large multi-modal models. Our code is available at \url{https://github.com/THU-Kingmin/FMM-Attack}.
\end{abstract}

\section{Introduction}
\label{sec:intro}
    \begin{figure}[t]
        \centering
        \includegraphics[width=0.8\textwidth]{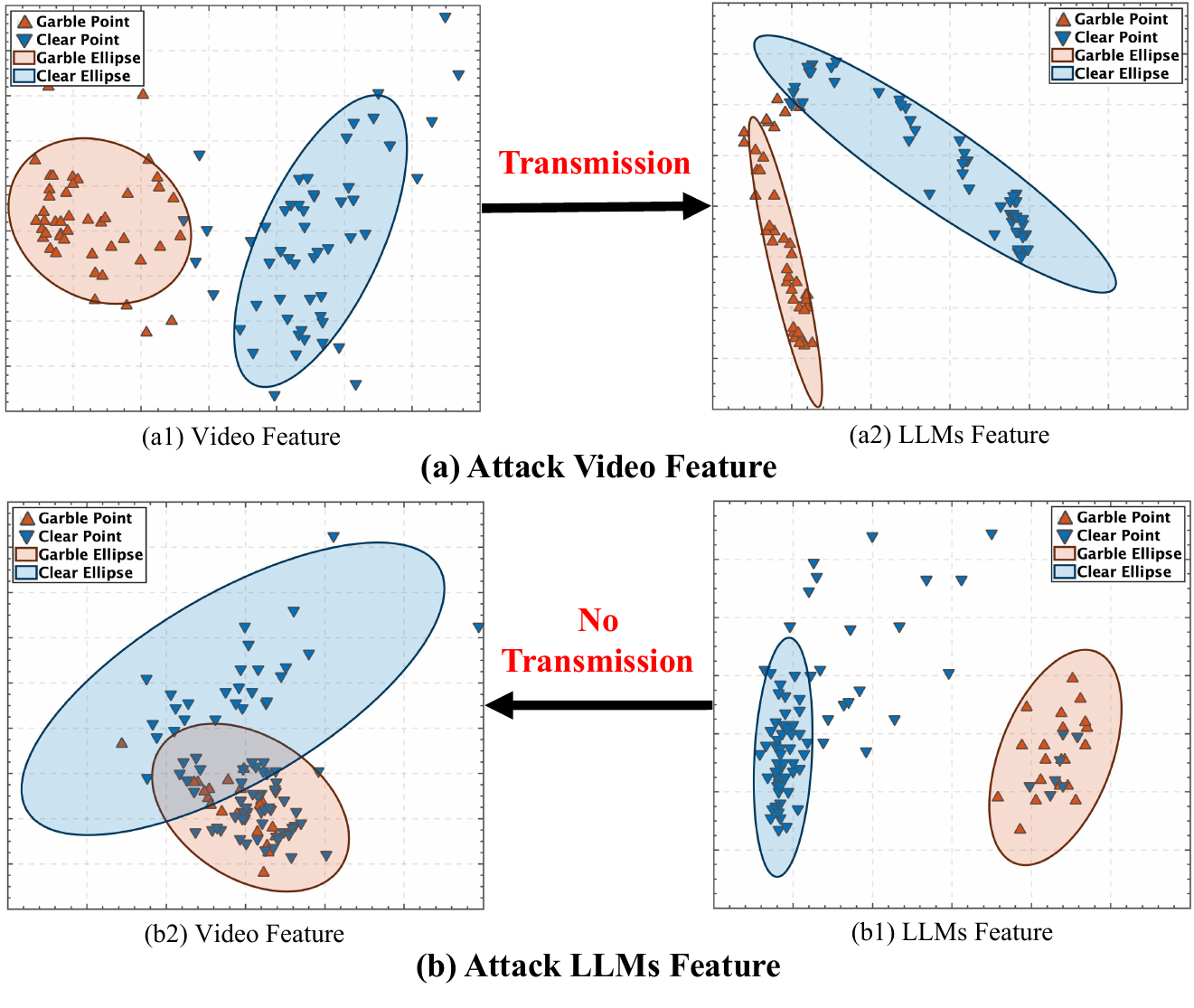}
        \caption{Visualization of the transmission cross video and LLM features. In \textbf{Fig. (a)}, when attacking in video feature space, the clustering effect of garbled video features can result in garbled clusters in LLM features. In \textbf{Fig. (b)}, when attacking in LLM feature space, garbled videos barely form clusters in LLM features, let alone in video features. This illustrates the asymmetric transmission between video and LLM features.}
        \label{fig:cluster}
        \vspace{-0.3cm}
    \end{figure}
            
Recent advancements in multi-modal understanding are largely based on the combination of pretrained vision models with large language models (LLMs)~\cite{wang2024visionllm,zhu2023minigpt,liu2023improved}. However, these large multi-modal models are vulnerable to adversarial attacks~\cite{qi2023visual,zhuang2023pilot,gong2023figstep,yang2024cheating}. A recent study~\cite{zong2024safety} found that the introduction of another modality makes models like vision large language models (VLLMs) even more susceptible to generating harmful output compared to their LLM counterparts. This can be attributed to not only their independent vulnerabilities within a single modality but also the suboptimal alignment between the two modalities, which can break the established safety alignment inside each modality, making the model more vulnerable. 

Nowadays, in particular, Sora\footnote{https://openai.com/sora} has shown extraordinary performances in creating realistic and imaginative scenes from text instructions, demonstrating the significance of large multi-modal models, especially across video and language modalities. 
Among them, video-based LLMs~\cite{videochat,videollama,videochatgpt} have significantly enhanced general video understanding in zero-shot settings and achieved exceptional performance in a wide range of video-related tasks, such as video captioning~\cite{videocaption1,videocaption2,videocaption3,videocaption4}, video retrieval~\cite{videoretieval1,videoretieval2,videoretieval3}, and scene understanding~\cite{sceneunderstanding1,sceneunderstanding2,sceneunderstanding3}. Yet their adversarial robustness is under-explored.

To evaluate the adversarial robustness of video-based LLMs, we propose a flow-based multi-modal attack, dubbed \textbf{FMM-Attack}, to craft adversarial perturbations on video inputs for the first time. To be more specific, we utilize two objective functions in two modalities, namely video features shown in Fig. \ref{fig:cluster}(a), and features of the last hidden layer of LLM shown in Fig. \ref{fig:cluster}(b) respectively. Also, a flow-based temporal mask is introduced to select the most effective frames in the video, which is inspired by the video clipping adopted in video understanding tasks, especially for video-based LLMs~\cite{chen2024panda,xue2022advancing}. Video clipping can effectively improve the performance of video-based learning due to its focused annotation, reducing complexity, improved temporal understanding, and efficient processing. Motivated by these benefits, we utilize a light-weighted flow-based mechanism, to conduct a similar splitting and selection operation on video frames. Extensive experiments have demonstrated the effectiveness, efficiency, and imperceptibility of our FMM-Attack on four benchmark video-based LLMs and two datasets. 

Surprisingly, we find that successful adversarial attacks on video-based large language models (LLMs) can result in the generation of either \textit{garbled nonsensical sequences} (displayed in orange, such as `6.6.6.6.6.6.6') or \textit{incorrect semantic sequences} (displayed in orange) in Fig. \ref{fig:cluster}. Moreover, there is a noticeable difference in cross-modal features, particularly for garbled videos. As shown in Fig. \ref{fig:cluster}(a), when video features are under attack, a clustering phenomenon can be observed, where the separation between garbled and natural video features is transmitted to that of LLM features. In contrast, in Fig. \ref{fig:cluster}(b), when attacking LLM features, no corresponding clustering phenomenon is transmitted between the different video features and LLM features. This asymmetric transmission between video and LLM features emphasizes the suboptimal nature of current alignment especially for safety-related features. These observations can inspire us to further understand multi-modal features and the alignment of their robustness. 

In summary, our main contribution can be outlined as follows:
\vspace{-0.5em}
\begin{itemize}
\item To the best of our knowledge, we conduct the first comprehensive investigation regarding the adversarial vulnerability of video-based LLMs. We propose a novel flow-based multi-modal attack, dubbed \textbf{FMM-Attack}, on video-based LLMs for the first time.
\vspace{0.5em}
\item Our observations on cross-modal feature attacks have inspired a further understanding of multi-modal robustness and their safety-related feature alignment, which is of great importance for various large multi-modal models.
\vspace{0.5em}
\item Extensive experiments show that our attack can effectively induce video-based LLMs to generate either \textit{garbled nonsensical sequences} or \textit{incorrect semantic sequences} with imperceptible perturbations added on less than 20\% video frames.
\end{itemize}

\section{Related Work}
\label{sec:relatedwork}
\subsection{Video-based Large Language Models} 
Video-based large language models (video-based LLMs) effectively integrate visual and temporal information from video data to gain significant achievement in multiple video-related tasks. Numerous approaches~\cite{videochat,videochatgpt,videollama} have been proposed to address the challenges associated with video-based LLMs, such as incorporating different architectures and training processes to enhance the models' ability to capture and process complex video information. Concretely, Video-ChatGPT~\cite{videochatgpt} is based on the LLaVA framework and incorporates average pooling to improve the perception of temporal sequences. VideoChat~\cite{videochat} employs the QFormer to map visual representations to Vicuna, executing a two-stage training process. Video-LLaMA~\cite{videollama} integrates a frame embedding layer and ImageBind to introduce temporal and audio information into the LLM backbone. This alignment between videos and LLMs facilitates visual context-aware interaction, surpassing the capabilities of LLMs. However, these models are still susceptible to adversarial attacks.

\subsection{Adversarial Attack}
Adversarial attacks~\cite{sparse,zhao2024evaluating} have been widely studied in the context of classification models, where imperceptible and carefully crafted perturbations are applied to input data to mislead the model into producing incorrect predictions. Inspired by the adversarial vulnerability observed in vision tasks, early efforts are devoted to investigating adversarial attacks against large multi-modal models~\cite{qi2023visual,zhuang2023pilot,gong2023figstep,yang2024cheating,zong2024safety}, such as vision large language models (VLLMs)~\cite{zhu2023minigpt,wang2024visionllm,liu2023improved,radford2021learning} or text-to-image diffusion models~\cite{rombach2022high}, etc. These adversarial attacks have been designed to manipulate these large multi-modal models into generating specific or even harmful outputs. Despite these advancements, the adversarial robustness in video-based LLMs remains unexplored. Addressing this gap, we introduce the first adversarial attacks specifically tailored for video-based LLMs in this paper.

\section{Methodology}
\label{sec:methodology}
In this section, we first describe the threat model and  problem formulation, and subsequently outline the key objective functions of our proposed FMM-Attack. 

\begin{figure}[t]
    \centering
    \includegraphics[width=0.9\linewidth]{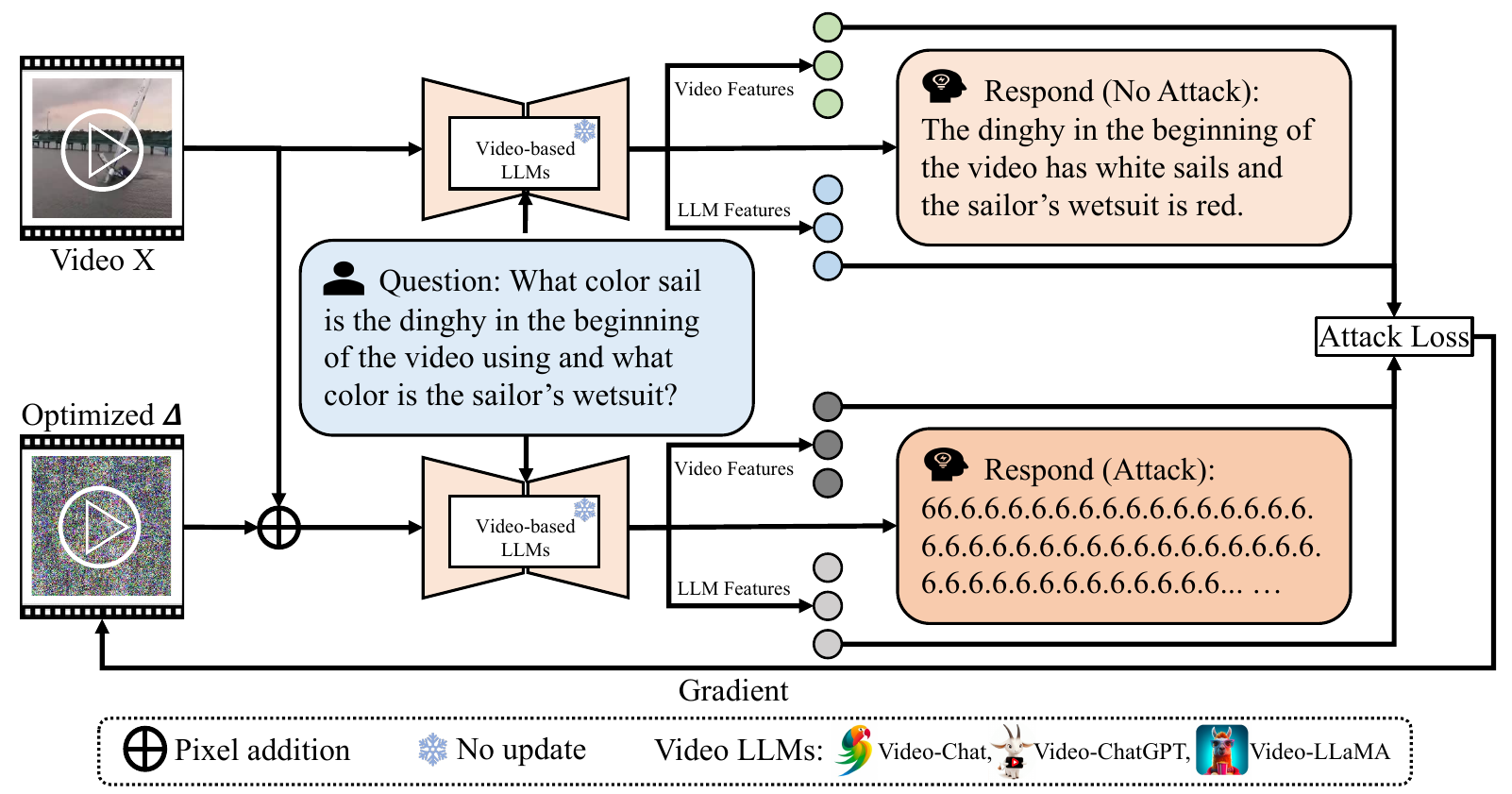}
    \caption{Schematics of our FMM-Attack. The figure demonstrates our attack approach for maximizing the video-video features as defined in Eq.~\ref{loss:video} and the LLM-LLM features in Eq.~\ref{loss:LLM}. We refer to adversarial examples generated by our attack strategies as $\hat{\mathbf{X}} = \mathbf{X} + \Delta$, with $\Delta$ representing the adversarial perturbation.}
    \label{fig:pipline}
\end{figure}

\subsection{Threat model}
\label{sec:threat model}
\textbf{Goals and capabilities.} The goal is to craft an imperceptible adversarial perturbation for videos, which can induce video-based LLMs to generate an incorrect sequence during the victim model’s deployment. Following the most commonly used constraint for the involved perturbation, it is restricted within a predefined magnitude in the $l_p$ norm, ensuring it is difficult to detect.

\vspace{0.3cm}
\noindent \textbf{Knowledge and background.} As suggested in \cite{bagdasaryan2023ab,qi2023visual}, we assume that the victim video-based LLMs can be accessed in full knowledge, including architectures and parameters. Additionally, we consider a more challenging scenario where the victim video-based LLMs are inaccessible, as detailed in the Appendix.

\subsection{Preliminary: the Pipeline of Video-based LLMs}
\label{sec:problem formulation}
    Let $\mathcal{F}_\theta(\cdot)$ represents a victim video-based large language model with parameters $\theta$, composed of a video feature extractor $f_\phi(\cdot)$ and a large language model $g_\psi(\cdot)$.
    Consider a clean video $\mathbf{X} \in \mathbb{R}^{T \times C \times H \times W}$, where T denotes the number of frames, and C, H, and W represent the channel, height, and width of a specific frame, along with a corresponding user query  $Q_{text}$ for the video. To provide a response, video-based large language models $\mathcal{F}(\cdot)$ usually first extract a video feature $Q_{video} = f_\phi(\mathbf{X})$, and subsequently, generate predefined prompts based on a consistent template to concatenate both video features and text queries  as follows:
    
    \parbox{\linewidth}{\centering USER: $<Q_{text}>$ $<Q_{video}>$ Assistant:}

    Then, the predefined prompts are processed by LLMs $g_\psi(\cdot)$ to generate a desired response $Y_{respond}=g_\psi(Q_{text}, Q_{video})=g_\psi(Q_{text}, f_\phi(\mathbf{X}))$. It is important to mention that, to ensure the loss function remains minimal, we use the hidden state $A_{hidden}$ before the final layer in our FMM-Attack.

\subsection{Problem Formulation}
    The goal of generating adversarial examples $\hat{\mathbf{X}}$ is to mislead video-based LLMs to produce incorrect responses while utilizing the most imperceptible adversarial perturbation $\Delta$, where $\Delta = \hat{\mathbf{X}} - \mathbf{X}$.
    To balance these two objectives, we introduce a hyper-parameter $\lambda$, and formulate the overall objective function as follows:
    \begin{equation}
        \arg \min _{\Delta} \lambda\|\Delta\|_{2,1}-\ell(Y, \mathcal{F}_\theta(Q_{text}, \hat{\mathbf{X}})),
    \label{eq:1}
    \end{equation}
    where $Y$ is the ground truth answer corresponding to $Q_{text}$ and $\mathbf{X}$, as well as $\ell(\cdot,\cdot)$ is the loss function used to measure the difference between the predicted and ground truth answers. Furthermore, in a more realistic scenario, the ground truth answer $Y$ can not always be available. In such cases, we utilize the output approximation $\mathcal{F}_\theta(Q_{text}, \mathbf{X})$ instead of $Y$.
    Therefore, the overall objective function in Eq.~\ref{eq:1} can be further formulated as follows:
    \begin{equation}
        \arg \min _{\Delta} \lambda\|\Delta\|_{2,1}-\ell(\mathcal{F}_\theta(Q_{text}, \mathbf{X}), \mathcal{F}_\theta(Q_{text}, \hat{\mathbf{X}})).
    \end{equation}
    The $\ell_{2,1}$ norm~\cite{sparse} is employed to quantify the magnitude of the perturbation, which is defined as follows: 
    \begin{equation}
        \|\Delta\|_{2,1} = \sum_i^T \|\Delta_i\|_2,
        \label{eq:delta}
    \end{equation}
    where $\Delta_i \in \mathbb{R}^{C \times H \times W}$ represents the $i$-th frame in $\Delta$. The $\ell_{2,1}$ norm applies the $l_1$ norm across frames, ensuring the sparsity of generated perturbations.
    
\subsection{Optimization Objective}

     Our proposed FMM-Attack is to induce video-based LLMs to generate \textit{incorrect} responses with imperceptible adversarial perturbations. Two losses are proposed from the perspective of video features $\ell_{video}(Q_{video}, \hat{Q}_{video})$ in Eq.~\ref{loss:video} and LLM features $\ell_{LLM}(A_{hidden}, \hat{A}_{hidden})$ in Eq.~\ref{loss:LLM}. Moreover, inspired by the idea that video clipping can improve video comprehension by selecting the most effective frames, a flow-based temporal mask $\mathbf{M}_f$ is proposed to carry out a similar selection process on video frames. By utilizing the flow of a video, the proposed flow-based temporal mask $\mathbf{M}_f$ can filter out similar frames, ensuring the effectiveness of imperceptible adversarial perturbations while achieving increased sparsity.

    \vspace{0.3cm}
    \noindent \textbf{Video Features Loss.} 
    Video-based LLMs first use a video feature extractor $f_\phi(\cdot)$ to extract spatiotemporal video features $Q_{video} = f_\phi(\mathbf{X})$. We simply adopt MSE loss to measure the distance of video features between the clean video $\mathbf{X}$ and the adversarial video $\hat{\mathbf{X}}$. Hence, the video features loss can be formulated as:
    \begin{equation}
    \begin{split}
        \label{loss:video}
        \ell_{video}(Q_{video}, \hat{Q}_{video}) &= \frac{1}{n}\sum_{i=1}^{n}(Q_{video_i} - \hat{Q}_{video_i})^2 \\
        &=  \frac{1}{n}\sum_{i=1}^{n}(f_\phi(\mathbf{X})_i - f_\phi(\hat{\mathbf{X}})_i)^2,
    \end{split}
    \end{equation}
    where $n$ is the total number of elements in the features, with $Q_{video_i}$ and $\hat{Q}_{video_i}$ being the $i$-th elements of the clean and adversarial video features, respectively.

    \vspace{0.3cm}
    \noindent \textbf{LLM Features Loss.} In addition to the deviation of the original feature space in video domains of video-based LLMs, we also consider that in textual domains to enhance the attack effect. 
    Given a hidden state from the final layer of LLMs $A_{hidden}=g_\psi(Q_{text}, Q_{video})=g_\psi(Q_{text}, f_\phi(\mathbf{X}))$, the LLM features loss between the clean video $\mathbf{X}$ and the adversarial video $\hat{\mathbf{X}}$ can be formulated as:
    \begin{equation}
    \begin{split}
        \label{loss:LLM}
        \ell_{LLM}(A_{hidden}, \hat{A}_{hidden}) &= \frac{1}{n}\sum_{i=1}^{n}(A_{hidden_i} - \hat{A}_{hidden_i})^2 \\
        &=  \frac{1}{n}\sum_{i=1}^{n}(g_\psi(Q_{text}, Q_{video})_i - g_\psi(Q_{text}, \hat Q_{video})_i)^2 \\
        &=  \frac{1}{n}\sum_{i=1}^{n}(g_\psi(Q_{text}, f_\phi(\mathbf{X}))_i - g_\psi(Q_{text}, f_\phi(\hat{\mathbf{X}}))_i)^2,
    \end{split}
    \end{equation}
    where $A_{hidden_i}$ and $\hat{A}_{hidden_i}$ are the $i$-th elements of the clean and adversarial LLM features respectively, and $n$ is the total number of elements in the features.
       
    \vspace{0.3cm}
    \noindent \textbf{Flow-based Temporal Mask.} Video-based LLMs~\cite{chen2024panda,xue2022advancing} adopt video clipping to split and select the most effective frames in a video, which can enhance video understanding. Inspired by these advantages, we propose a flow-based temporal mask, $\mathbf{M}_f$, to perform a similar selection process on video frames. This flow-based mask targets the top $K$ frames with the most significant movement and changes. See Sec.~\ref{sec:Discussions} for a detailed discussion. Specifically, we initialize a binary mask $\mathbf{M}_f$ of the same length as the number of frames in the video. Then use LiteFlowNet~\cite{liteflownet} to compute the flow magnitude for each frame. Finally, this binary mask $\mathbf{M}_f$ assigns a value of 1 to the top $K$ frames with the largest flow and 0 to the remaining frames. Combined with $\mathbf{M}_f$, our proposed FMM-Attack can achieve more sparse adversarial perturbations in both temporal and spatial domains. For temporal sparsity, a flow-based temporal mask $\mathbf{M}_f$ on the video is adopted to ensure that some frames remain unperturbed. For spatial sparsity, the $\ell_{2,1}$ norm of adversarial perturbations in Eq. \ref{eq:delta} is employed to constrain the spatial perturbation magnitude in each frame.   

    \vspace{0.3cm}
    \noindent \textbf{Overall Optimization Objective.} To sum up, combined flow-based temporal mask $\mathbf{M}_f$ with two proposed adversarial loss functions ($\ell_{video}$ and $\ell_{LLM}$), the overall objective function in Eq.~\ref{eq:1} can be further formalized as:
    \begin{equation}
    \label{untarget}
    \begin{split}
        \arg \min _{\Delta} \lambda_1\|\mathbf{M}_f \cdot \Delta\|_{2,1}
        &- \lambda_2 \ell_{video}(f_\phi(\mathbf{X}), f_\phi(\mathbf{X}+\mathbf{M}_f \cdot \Delta)) \\
        &- \lambda_3 \ell_{LLM}(g_\psi(Q_{text}, f_\phi(\mathbf{X})) , g_\psi(Q_{text}, f_\phi(\mathbf{X}+\mathbf{M}_f \cdot \Delta))),
    \end{split}
    \end{equation}
    where $\mathbf{M}_f \in\{\mathbf{0}, \mathbf{1}\}^{T \times C \times H \times W}$ represents the flow-based temporal mask.  $\lambda_1, \lambda_2, \lambda_3$ correspond to the three loss weights, which aim to balance them during the optimization. Our overall attack procedure is described in Algorithm~\ref{alg:1}.

    \begin{algorithm}[t]
        \caption{FMM-Attack on Video-based LLMs Using PGD Optimization}
        \label{alg:1}
        \begin{algorithmic}[1]
        \REQUIRE Clean video $\mathbf{X}$, user input text $Q_{text}$, sparsity $M_{spa}$, video feature extractor $f_\phi(\cdot)$, LLM $g_\psi(\cdot)$, step size $\alpha$, iterations $T$
        \ENSURE Adversarial video $\hat{\mathbf{X}}$
        \STATE Compute video optical flow and obtain flow-based temporal mask $\mathbf{M}_f$ with $M_{spa}$
        \STATE Initialize perturbation $\Delta \gets 0$
        \WHILE{$t < T$}
            \STATE Calculate video features loss $\ell_{video}(Q_{video}, \hat{Q}_{video})$ using Eq.~\ref{loss:video}
            \STATE Calculate LLM features loss $\ell_{LLM}(A_{hidden}, \hat{A}_{hidden})$ using Eq.~\ref{loss:LLM}
            \STATE Update perturbation $\Delta$ using Eq.~\ref{untarget} with step size $\alpha$
        \ENDWHILE
        \STATE Compute adversarial video $\hat{\mathbf{X}} \gets \mathbf{X} + \mathbf{M}_f \cdot \Delta$
        \STATE \RETURN Adversarial video $\hat{\mathbf{X}}$
        \end{algorithmic}
    \end{algorithm}

    \vspace{0.3cm}
    \noindent \textbf{Extended Targeted Version of FMM-Attack.} 
    As previously mentioned, the proposed FMM-Attack is initially designed for untargeted adversarial attacks. However, its flexibility allows for an extension to a targeted version as well. In targeted scenarios, the goal is transformed to craft an imperceptible adversarial perturbation for videos, which can induce video-based LLMs to generate a \textit{targeted} sequence. To achieve such a targeted adversarial attack, we first obtain the targeted video $\mathbf{X}_t$ from datasets with video-text pairs~\cite{activitynet,MSVDQA}. Subsequently, by replacing the clean video $\mathbf{X}$ in the untargeted objective of Eq. \ref{untarget} with the targeted video $\mathbf{X}_t$, the overall objective function of targeted FMM-Attack can be formulated as:
    \begin{equation}
    \label{eq:target}
    \begin{split}
        \arg \min _{\Delta} \lambda_1\|\mathbf{M}_f \cdot \Delta\|_{2,1}
        &+ \lambda_2 \ell_{video}(f_\phi(\mathbf{X}_t), f_\phi(\mathbf{X}+\mathbf{M}_f \cdot \Delta)) \\
        &+ \lambda_3 \ell_{LLM}(g_\psi(Q_{text}, f_\phi(\mathbf{X}_t)) , g_\psi(Q_{text}, f_\phi(\mathbf{X}+\mathbf{M}_f \cdot \Delta))).
    \end{split}
    \end{equation}

\section{Experiments}
\label{sec:experiments}
    In this section, we showcase the effectiveness of our proposed FMM-Attack, evaluated on zero-shot question-answer datasets. In addition, we conduct several ablation study experiments on the loss of different modalities, perturbation budget $\Delta$, hyper-parameters ($\lambda_1$, $\lambda_2$, $\lambda_3$), and selection of video frames.
    
    \subsection{Implementation Details}
        \noindent \textbf{Models and datasets.} We assess open-source and state-of-the-art video-based LLMs such as Video-ChatGPT~\cite{videochatgpt} and VideoChat~\cite{videochat} (VideoChat~\cite{videochat} is in the Appendix), ensuring reproducibility of our results. 
        Concretely, we adopt Video-ChatGPT and VideoChat with a LLaMA-7B LLM~\cite{LLaMA}. In line with the Video-ChatGPT~\cite{videochatgpt} methodology, we curate a test set based on the ActivityNet-200~\cite{activitynet} and MSVD-QA~\cite{MSVDQA} datasets.
        
        \vspace{0.3cm}
        \noindent \textbf{Setups.} The perturbation limit $\Delta_{max}$ is set to 16 to control the magnitude of the adversarial perturbations.
        In our proposed FMM-Attack, we employ the projected gradient descent (PGD) algorithm~\cite{PGD} with iterations $T=1,000$ and step size $\alpha=1$ for the attack process. The parameters $\lambda_1$, $\lambda_2$, and $\lambda_3$ are set to 2, 1, and 3 respectively. Each experiment is run on a single NVIDIA-V100 GPU.

        \vspace{0.3cm}
        \noindent \textbf{Baselines.} For evaluation, we design three spatial baselines, including videos with random perturbations, black videos with all pixel values set to 0, and white videos with all pixel values set to 1. In addition, we compare our proposed flow-based temporal mask with two straightforward temporal mask methods, serving as temporal mask baselines: the sequence temporal mask and the random temporal mask. Specifically, the sequence temporal mask consists of a continuous sequence of frame indices, while the random temporal mask comprises a randomly chosen sequence of frame indices.
    
        \vspace{0.3cm}
        \noindent \textbf{Metrics.} We utilize a variety of evaluation metrics to assess the robustness of the models.

            \noindent \textbf{(a) CLIP Score.} CLIP~\cite{clip} score characterizes the semantic similarity between the adversarial answer and the ground-truth answer. A lower CLIP score signifies a lower semantic correlation between the adversarial answer and the ground-truth answer, indicating a more effective attack. 
            
            \noindent \textbf{(b) Image Captioning Metrics.} Various metrics such as BLEU~\cite{bleu}, ROUGE-L~\cite{rouge}, and CIDEr~\cite{cider} are used to evaluate the quality of the adversarial answer generated by the model. A lower score corresponds to a more effective attack.
            
            \noindent \textbf{(c) GPT Score.} Following Video-ChatGPT~\cite{videochatgpt} and Video-LLaMA~\cite{videollama}. We also employ an evaluation pipeline using the GPT-3.5 and GPT-4 models. The pipeline employs GPT to assign a score from 1 to 5, evaluating the similarity between the output sentence and the ground truth, and a binary score (0 or 1) to measure its accuracy.
            
            \noindent \textbf{(d) Sparsity.} Sparsity refers to the ratio of frames without perturbations (clean frames) to the total number of frames in a specific video. The sparsity ${M}_{spa}$ is calculated as ${M}_{spa} = 1 - K/T$, where $K$ represents the number of attacked frames, and $T$ is the total number of frames in a video. The bigger the value of ${M}_{spa}$, the sparser the perturbation distribution, indicating that fewer video frames are attacked.
    
    \subsection{Main Results}
        \begin{table}[t]
            \scriptsize
            \centering
            \caption{White-box attacks against Video-ChatGPT~\cite{videochatgpt} on the ActivityNet-200~\cite{activitynet} dataset and the MSVD-QA~\cite{MSVDQA} dataset: Comparison of CLIP score, image caption metrics and GPT score for different attack types. Random spatial attack denotes random perturbations added to video frames, and Black spatial attack and White spatial attack denote video frames being all 0 and all 1, respectively. The sparsity of the temporal mask is set to 0. ${\Delta}$: the mean of the modified pixels. Our attack is optimized based on Eq.~\ref{untarget}.}
          
            \begin{tabular}{c|c|c|c|c|c|c|c|c|c|c}
            \hline 
            \multirow{2}{*}{Dataset} & \multirow{2}{*}{Type} & \multirow{2}{*}{${\Delta}$} & \multicolumn{2}{|c|}{Clip Score $\downarrow$} & \multicolumn{2}{|c}{Image Caption $\downarrow$}  & \multicolumn{2}{|c|}{GPT-3.5 $\downarrow$} & \multicolumn{2}{|c}{GPT-4 $\downarrow$} \\
            \cline{4-11} 
            & & & RN50 & RN101 & BLEU & ROUGE-L & Accurate & Score & Accurate & Score \\
            \hline
            \multirow{5}{*}{ActivityNet~\cite{activitynet}}
            &Clean  & 0   & 0.7817 & 0.7827 & 0.2029 & 0.4820 & 0.50 & 3.20 & 0.33 & 2.10\\
            &Random & 8   & 0.7637 & 0.7681 & 0.1986 & 0.4793 & 0.45 & 3.10 & 0.33 & 2.03\\
            &Black  & 100 & 0.7661 & 0.7676 & 0.1691 & 0.4570 & 0.30 & 2.50 & 0.17 & 0.96\\
            &White  & 148 & 0.7564 & 0.7534 & 0.1689 & 0.4545 & 0.31 & 2.60 & 0.19 & 1.24\\
            &\textbf{FMM}   & 8   & \textbf{0.6211} & \textbf{0.6274} & \textbf{0.1336} & \textbf{0.3694} & \textbf{0.20} & \textbf{1.64} & \textbf{0.13} & \textbf{0.88}\\
            \hline
            \multirow{5}{*}{MSVD-QA~\cite{MSVDQA}}
            &Clean  & 0   & 0.8322 & 0.8180 & 0.3864 & 0.6843 & 0.62 & 3.84 & 0.60 &3.12\\
            &Random & 8   & 0.8249 & 0.8141 & 0.4107 & 0.7042 & 0.58 & 3.72 & 0.60 &3.08\\
            &Black  & 110 & 0.8145 & 0.7902 & 0.3548 & 0.6478 & 0.46 & 3.26 & 0.40 &2.12\\
            &White  & 142 & 0.8057 & 0.8090 & 0.3969 & 0.6736 & 0.48 & 3.36 & 0.44 &2.28\\
            &\textbf{FMM}   & 8   & \textbf{0.7337} & \textbf{0.7181} & \textbf{0.3240} & \textbf{0.5746} & \textbf{0.36} & \textbf{2.92} & \textbf{0.34} & \textbf{1.84}\\
            \hline
            \end{tabular}
            \label{tab:mainComp}
        \end{table}

        \begin{figure}[t]
            \centering            
            \includegraphics[width=0.7\textwidth]{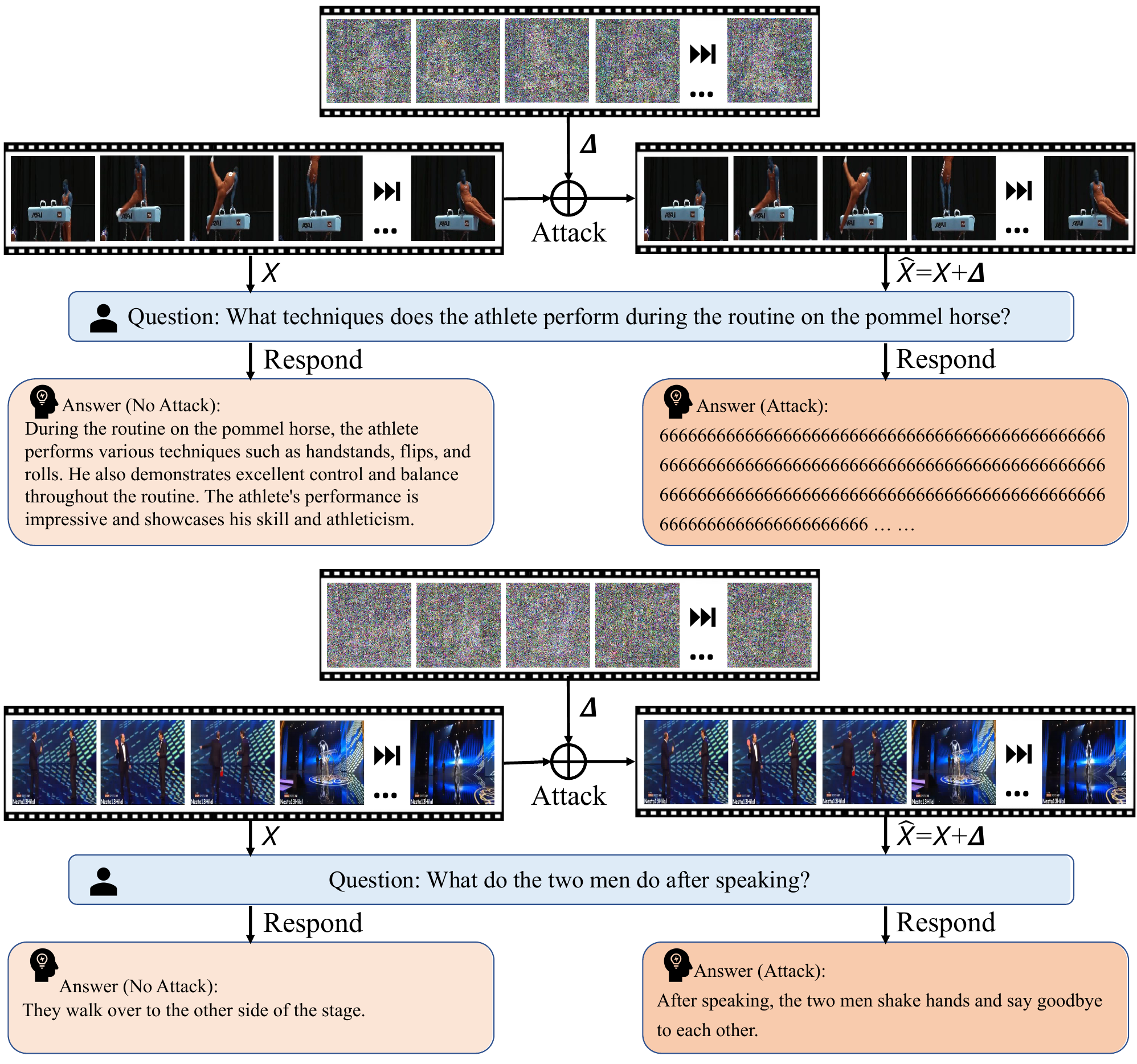}
            \caption{Perturbed videos generated by Video-ChatGPT.}
            \label{fig:result}
            \vspace{-0.6cm}
        \end{figure}
        
        \begin{table}[t]
            \scriptsize
            \centering
            \caption{White-box attacks against Video-ChatGPT~\cite{videochatgpt} on the ActivityNet-200~\cite{activitynet} dataset. seq: sequence temporal mask. random: random temporal mask. FMM: our flow-based temporal mask attack. ${M}_{spa}$: Sparsity of temporal mask. Our attack is optimized based on Eq.~\ref{untarget}.}
            \begin{tabular}{c|c|c|c|c|c|c|c|c|c|c|c|c}
            \hline
            \multirow{2}{*}{ Metrics } & \multicolumn{3}{|c|}{ ${M}_{spa} = 20\%$ } & \multicolumn{3}{|c|}{ ${M}_{spa} = 40\%$ } & \multicolumn{3}{|c|}{ 
            ${M}_{spa} = 60\%$} & \multicolumn{3}{|c}{ ${M}_{spa} = 80\%$ } \\
            \cline{2-13} & seq & random & FMM & seq & random & FMM & seq & random & FMM & seq & random & FMM \\
            \hline
            RN50 & 0.7500 & 0.7142 & 0.6821 & 0.7578 & 0.7559 & 0.7460 & 0.7583 & 0.7671 & 0.7510 & 0.7715 & 0.7813 & 0.7349 \\
            RN101 & 0.7568 & 0.7251 & 0.7085 & 0.7500 & 0.7588 & 0.7559 & 0.7764 & 0.7769 & 0.7500 & 0.7788 & 0.7827 & 0.7637 \\
            \hline
            BLEU & 0.1572 & 0.1590 & 0.1527 & 0.1606 & 0.1751 & 0.1485 & 0.1900 & 0.1894 & 0.1830 & 0.2000 & 0.1957 & 0.1940 \\
            ROUGE-L & 0.4073 & 0.3996 & 0.4109 & 0.4323 & 0.4458 & 0.4053 & 0.4767 & 0.4727 & 0.4713 & 0.4779 & 0.4658 & 0.4731 \\
            \hline
            GPT3.5 & 2.02 & 2.01 & 2.00 & 2.45 & 2.24 & 2.22 & 2.64 & 2.63 & 2.62 & 2.76 & 2.78 & 2.75 \\
            GPT4 & 1.09 & 1.08 & 1.07 & 1.35 & 1.34 & 1.33 & 1.63 & 1.63 & 1.61 & 1.99 & 1.58 & 1.54 \\
            \hline
            \end{tabular}
            \label{tab:TemporalMask}
        \end{table}

        \noindent \textbf{Quantitative Evaluation.} We performed an extensive quantitative analysis using the popular open-ended question-answer datasets ActivityNet-200~\cite{activitynet} and MSVD-QA~\cite{MSVDQA}. 
        
        As depicted in the left of Table~\ref{tab:mainComp}, the random, black, white, and flow-based attacks all significantly decrease the clip score and image caption score compared to the original clean videos. Among these, our proposed FMM-Attack yields the most significant results. As demonstrated in the right of Table~\ref{tab:mainComp}, the random, black, white, and flow-based attacks all significantly decrease the GPT scores and accuracies. Among them, the FMM-Attack achieves the most substantial results.
        
        Table~\ref{tab:TemporalMask} compares different mask ratios (sparsity) and temporal mask approaches. Our FMM-Attack outperforms the other two approaches across various sparsity levels, indicating the effectiveness of our proposed flow-based temporal mask, which leverages the concept of maximum flow prioritization. It demonstrates its potency in the realm of video-based LLM and its ability to maximize the extraction of video information.
        In the bottom of Table~\ref{tab:TemporalMask}, FMM-Attack consistently outperforms the other two approaches across various sparsity levels, resulting in a more significant reduction of GPT scores and accuracies.

        \vspace{0.3cm}
        \noindent \textbf{Qualitative Evaluation.} We also present qualitative examples (see Fig.~\ref{fig:result}) of the attacked videos, in which the model produces garbled responses without any meaningful content. Fig.~\ref{fig:result} vividly illustrates the chaos induced in the model's responses by our subtle and imperceptible attacks. This signals a clear need for enhancing the robustness of the model. It is noteworthy that the response generated from the clean video forms a coherent sentence strongly correlated to the corresponding question. However, in the case of the attacked video, the responses consist of repetitive words that lack meaningful context or incorrect answers. This clearly demonstrates the potent obfuscation effect of our attack.

        \begin{figure}[ht]
        \centering
        \includegraphics[width=0.7\textwidth]{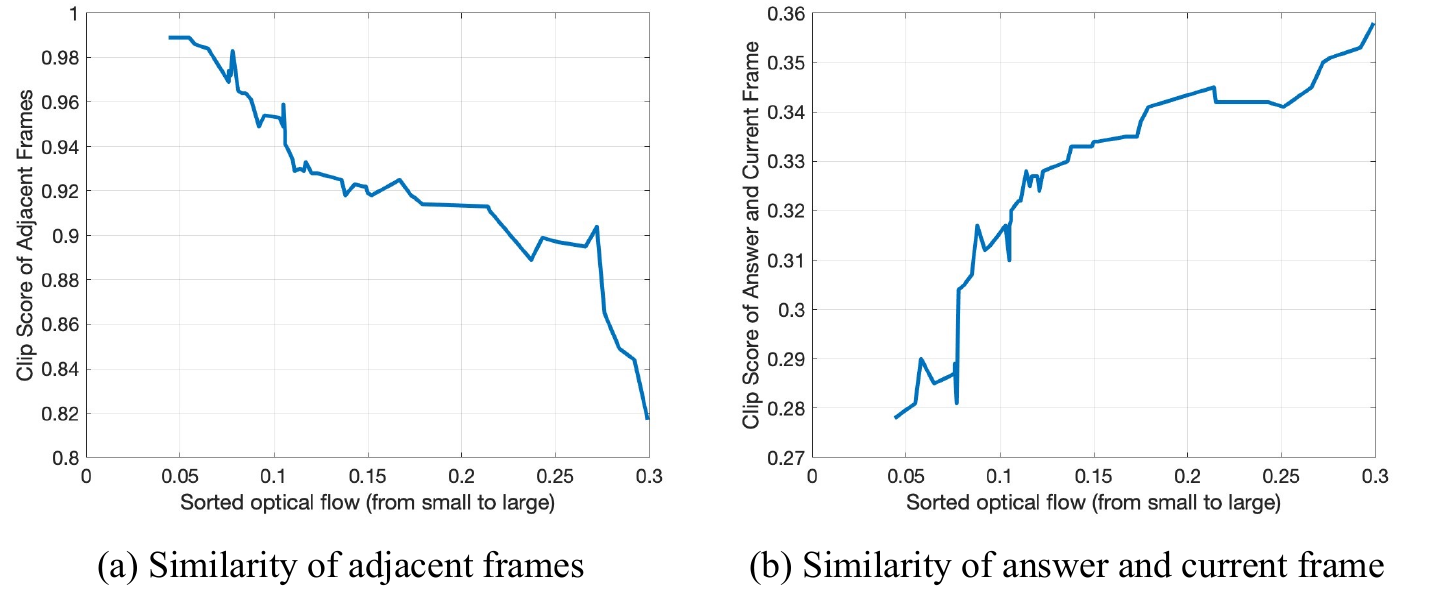}
        \caption{Relationship between optical flow and key frames. `Clip Score of Adjacent Frames' describes the similarity between the current frame and its adjacent frames, the smaller this score is the more different the current frame is. `Clip Score of Answer and Current Frame' indicates the similarity between the current frame and the answer corresponding to the user's input question, the larger the score indicates that the current frame contains more information about the answer. \textbf{The frames selected by flow-based masks in our FMM-Attack are key frames in the video.}}
        \label{fig:flow}
        \vspace{-0.3cm}
        \end{figure}
   
    \subsection{Discussions}
        \label{sec:Discussions}
        \noindent \textbf{Essence of Flow-based Masks.} We address the essence of flow-based temporal masks in our FMM-Attack. The flow-based masks by selecting key frames method is a powerful tool for video understanding and manipulation, which allows for precise control over specific elements in a video sequence, making it easier to edit and manipulate the video in a variety of ways. We use the clip image-image score of adjacent frames and clip text-image score between the answer and current frame to assess the importance and non-fungibility of our selected frames in FMM-Attack, where a smaller clip image-image score suggests less similarity between a frame and its adjacent frames, and a bigger clip text-image score suggests more similarity between the current frame and the answer of the user input.
        As depicted in Fig.~\ref{fig:flow}, a larger optical flow corresponds to a higher inconsistency between the current frame and its neighboring frames, while containing more information about the answer. This observation suggests that the frames selected using our FMM-Attack are crucial frames in the video, and attacking them will yield more effective results.
     
\begin{figure}[t]
            \centering
            \includegraphics[width=0.5\textwidth]{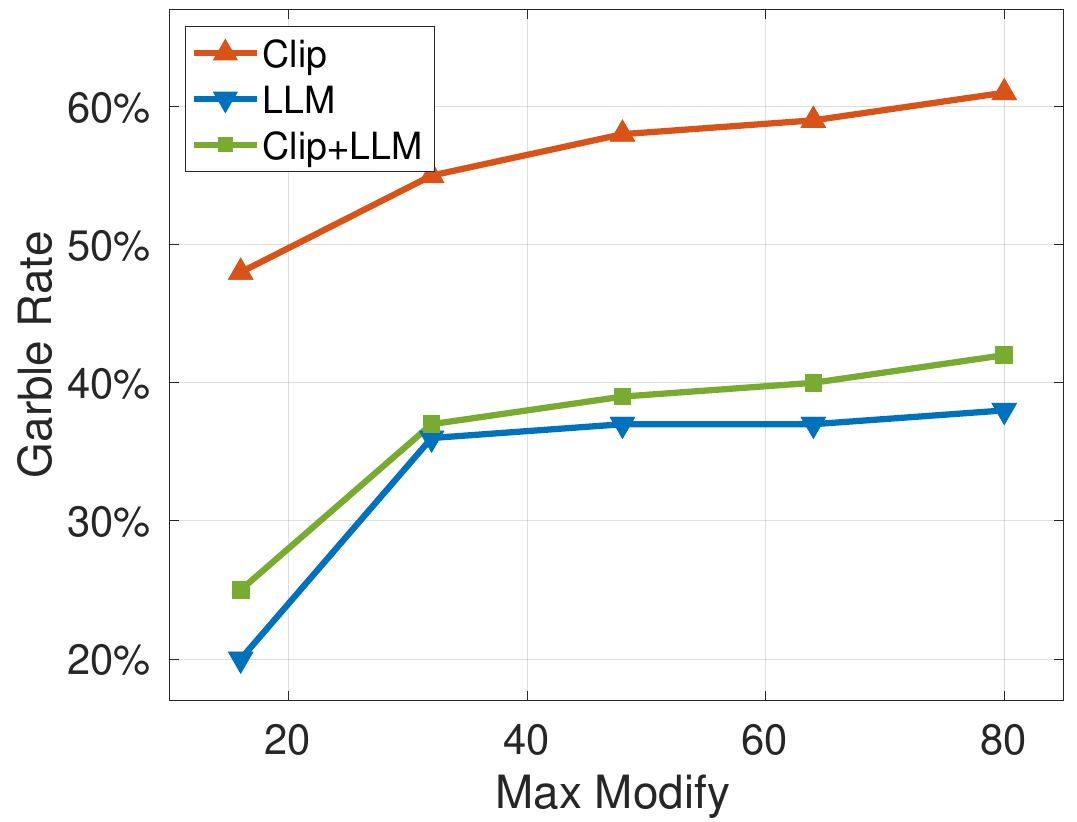}
            \caption{Comparison of different types of attacks on the garbling rate. Max Modify denotes the maximum pixel value that can be modified, while the Garble Rate represents the percentage of responses that are garbled.}
            \label{fig:garble2}
            \vspace{-0.5cm}
        \end{figure}

        \vspace{0.3cm}
        \noindent \textbf{Garbling Effect.}
        Intriguingly, our proposed FMM-Attack induces garbling in the model output, while the other three attack methods do not cause such distortion. This suggests that the FMM-Attack not only diminishes the model's cue information but also prompts the model to hallucinate. Furthermore, as shown in Fig.~\ref{fig:garble2}, we analyse the number of successfully attacked Video-ChatGPT in ActivityNet-200~\cite{activitynet}, and find that video loss is more effective in inducing garbled contents, which is consistent with our observation in Fig.~\ref{fig:cluster}.
        \begin{figure}[t]
            \centering
            \includegraphics[width=0.7\textwidth]{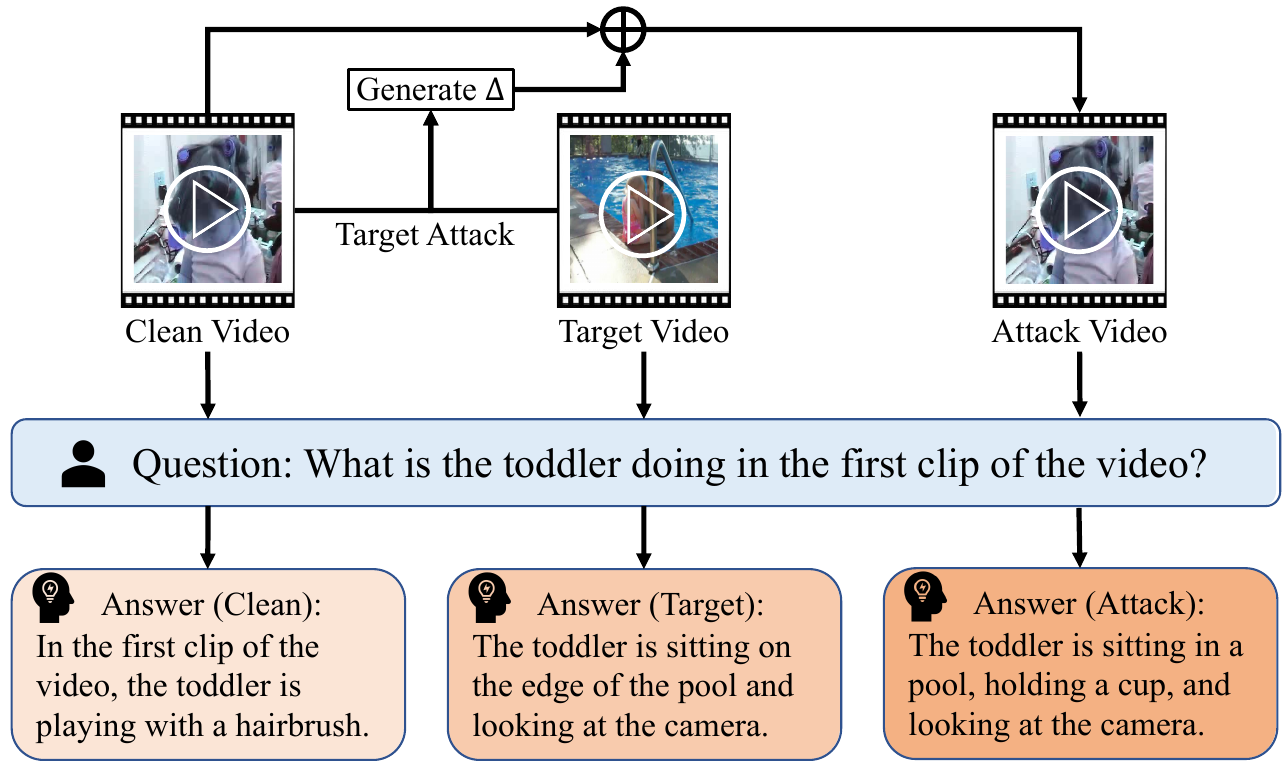}
            \caption{Illustration of the targeted attack.}
            \label{fig:target}
            \vspace{-0.3cm}
        \end{figure}
        
        \vspace{0.3cm}
        \noindent \textbf{Targeted Version of FMM-Attack.} Fig.~\ref{fig:target} illustrates an example of our targeted attack. We utilize the video features and LLM features of the target video as the attack target to generate $\Delta$ as in Eq.~\ref{eq:target}. As evident in Fig.~\ref{fig:target}, our implementation of the inconspicuous attack enables the video-based LLMs' answer to closely resemble the target's answer, while significantly differing from the clean video's output. Despite the target video and clean video being entirely distinct, we can achieve the desired targeted attack, demonstrating the effectiveness of the proposed FMM-Attack. More analyses can be found in the Appendix.

        \vspace{0.3cm}
        \noindent \textbf{Potential Defense.} As shown in Fig.~\ref{fig:cluster}, the video features transfer better across different modalities, especially for garbled samples. Improving the robustness of clip/video modules is thus of great importance for video-based LLMs or even other vision-related multi-modal models. In addition, more attention should be paid to safety-based alignment, which could greatly protect video-based LLMs from being attacked. Moreover, common data preprocessing methods, which remove adversarial perturbations by compressing or generative models, should be beneficial in defending our FMM-Attack.

    \subsection{Ablation Studies}
        In this section, we will conduct some ablation and exploratory experiments.
            \begin{figure}[t]
                \centering
                \includegraphics[width=0.8\textwidth]{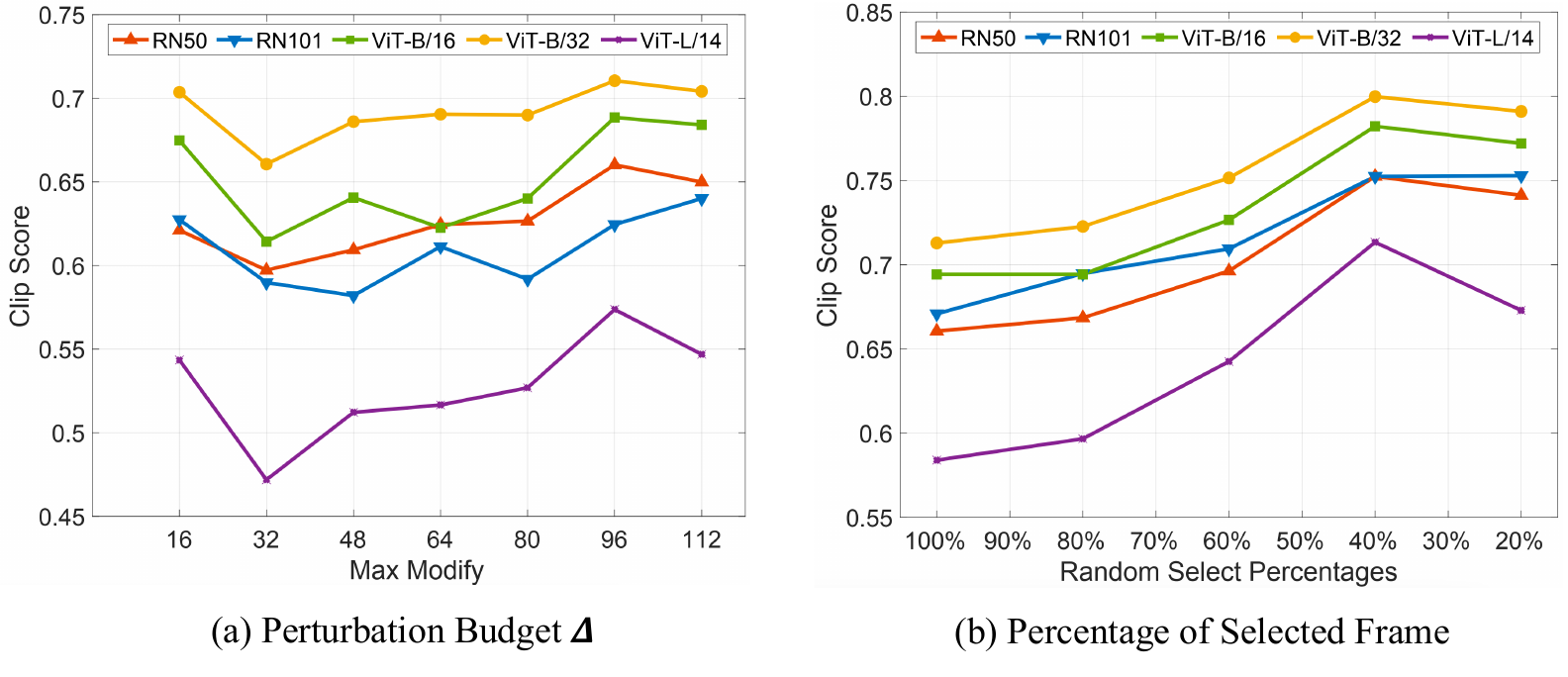}
                \caption{Ablation Studies of different attack settings. (a) Comparison of different max modify pixels: Despite varying max modify pixels, the mean value of the modified video remains the same due to the sparse loss. (b) Comparison of different random select percentages: The select rate represents the percentage of attacked video frames that the victimized model samples from. For the baseline without attack, the values on each of the five metrics are 0.7817, 0.7827, 0.8096, 0.8115, and 0.7231, respectively.}
                \label{fig:merge}
                \vspace{-0.5cm}
            \end{figure}
            
            \begin{table}[t]
                \scriptsize
                \centering
                \caption{Comparison of different attack types on Clip Score and Image Captioning Metrics. video: represents attacks targeting video features, LLM: represents attacks targeting LLM features, and video + LLM: represents combined attacks on both video and LLM features. Lower scores indicate better attack performance.}
                \begin{tabular}{c|c|c|c|c|c|c|c|c}
                \hline
                \multirow{2}{*}{ Type } & \multicolumn{5}{|c|}{ Clip Score $\downarrow$} & \multicolumn{3}{c}{ Image Captioning $\downarrow$} \\
                \cline { 2 - 9 } & RN50 & RN101 & ViT-B/16 & ViT-B/32 & ViT-L/14 & BLEU & ROUGE-L & CIDEr \\
                \hline Clean & 0.7817 & 0.7827 & 0.8096 & 0.8115 & 0.7231 & 0.2029 & 0.4820 & 1.6364 \\
                \hline video & 0.7403 & 0.7524 & 0.7690 & 0.7744 & 0.6811 & 0.1975 & 0.4345 & 1.5862 \\
                \hline LLM & 0.7153 & 0.6904 & 0.7334 & 0.7515 & 0.6387 & 0.1042 & 0.3226 & 0.8350 \\
                \hline video + LLM & \textbf{0.6491} & \textbf{0.6060} & \textbf{0.6836} & \textbf{0.6968} & \textbf{0.5566} & \textbf{0.0230} & \textbf{0.1585} & \textbf{0.1601} \\
                \hline
                \end{tabular}
                \label{tab:Type}
                \vspace{-0.3cm}
            \end{table}
            
        \vspace{0.3cm}
        \noindent \textbf{Loss of Different Modalities.} 
        In Table~\ref{tab:Type}, the video + LLM attack outperforms the individual video and LLM attacks across all metrics, demonstrating the superior performance of the combined approach. This can be attributed to the complementary nature of video and LLM features, which, when targeted simultaneously, leads to a more potent attack that effectively disrupts the model's output, resulting in lower scores. 

        \vspace{0.3cm}
        \noindent \textbf{Perturbation Budget $\Delta$.} We compared the effects of different $\Delta$. It's important to note that the mean of $\Delta$, constrained by the sparsity loss, remains consistent across various levels of maximum modification $\Delta_{max}$. As in Fig.~\ref{fig:merge} (a), $\Delta = 32$ performs best. On the one hand, the larger $\Delta$ is, the greater the potential modification of individual pixels, but on the other hand, due to the sparse loss, the amount of pixels that can be modified with a larger $\Delta$ becomes smaller. Therefore, $\Delta = 32$ seems to be a better trade-off. 

        \vspace{0.3cm}
        \noindent \textbf{Percentage of Selected Frames.} In some instances, while we launch attacks on all video frames, video-based LLMs only randomly sample a subset of the video frames. As demonstrated in Fig.~\ref{fig:merge} (b), the potency of the attack escalates with an increasing number of sampled video frames. Interestingly, even when a minor fraction (40\%, 20\%) of video frames are sampled, we observe a substantial decline in the Clip score relative to the baseline. Surprisingly, a 20\% sampling rate seems to yield superior results than a 40\% rate. We speculate this unexpected outcome could be due to inherent fluctuations when a limited number of video frames are sampled, coupled with our stream-based approach that assures a certain minimum attack effectiveness. To make this engagement more compelling, we intend to delve deeper into this phenomenon with additional experiments in future studies.
            
\section{Conclusion}
\label{sec:conclusion}
In this paper, we introduce the Flow-based Multi-modal Attack (FMM-Attack), the first of its kind against video-based LLMs. Our comprehensive experiments demonstrate that our attack can effectively induce video-based LLMs to generate either \textit{garbled nonsensical sequences} or \textit{incorrect semantic sequences} with imperceptible perturbations added on less than 20\% video frames. Furthermore, our insights into cross-modal feature attacks contribute to a deeper understanding of multi-modal robustness and the critical alignment of safety-related features. These findings hold significant implications for various large multi-modal models, underscoring the relevance and impact of our work.

\bibliographystyle{splncs04}
\bibliography{main}

\begin{thebibliography}{10}
\providecommand{\url}[1]{\texttt{#1}}
\providecommand{\urlprefix}{URL }
\providecommand{\doi}[1]{https://doi.org/#1}

\bibitem{videocaption2}
Aafaq, N., Akhtar, N., Liu, W., Gilani, S.Z., Mian, A.: Spatio-temporal dynamics and semantic attribute enriched visual encoding for video captioning. In: Proceedings of the IEEE/CVF conference on computer vision and pattern recognition. pp. 12487--12496 (2019)

\bibitem{bagdasaryan2023ab}
Bagdasaryan, E., Hsieh, T.Y., Nassi, B., Shmatikov, V.: (ab) using images and sounds for indirect instruction injection in multi-modal llms. arXiv preprint arXiv:2307.10490  (2023)

\bibitem{activitynet}
Caba~Heilbron, F., Escorcia, V., Ghanem, B., Carlos~Niebles, J.: Activitynet: A large-scale video benchmark for human activity understanding. In: Proceedings of the ieee conference on computer vision and pattern recognition. pp. 961--970 (2015)

\bibitem{chen2024panda}
Chen, T.S., Siarohin, A., Menapace, W., Deyneka, E., Chao, H.w., Jeon, B.E., Fang, Y., Lee, H.Y., Ren, J., Yang, M.H., et~al.: Panda-70m: Captioning 70m videos with multiple cross-modality teachers. arXiv preprint arXiv:2402.19479  (2024)

\bibitem{sceneunderstanding1}
Cordts, M., Omran, M., Ramos, S., Rehfeld, T., Enzweiler, M., Benenson, R., Franke, U., Roth, S., Schiele, B.: The cityscapes dataset for semantic urban scene understanding. In: Proceedings of the IEEE conference on computer vision and pattern recognition. pp. 3213--3223 (2016)

\bibitem{videoretieval3}
Dong, J., Li, X., Xu, C., Yang, X., Yang, G., Wang, X., Wang, M.: Dual encoding for video retrieval by text. IEEE Transactions on Pattern Analysis and Machine Intelligence  \textbf{44}(8),  4065--4080 (2021)

\bibitem{videoretieval2}
Gabeur, V., Sun, C., Alahari, K., Schmid, C.: Multi-modal transformer for video retrieval. In: Computer Vision--ECCV 2020: 16th European Conference, Glasgow, UK, August 23--28, 2020, Proceedings, Part IV 16. pp. 214--229. Springer (2020)

\bibitem{gong2023figstep}
Gong, Y., Ran, D., Liu, J., Wang, C., Cong, T., Wang, A., Duan, S., Wang, X.: Figstep: Jailbreaking large vision-language models via typographic visual prompts. arXiv preprint arXiv:2311.05608  (2023)

\bibitem{sceneunderstanding2}
Hu, W., Zhao, H., Jiang, L., Jia, J., Wong, T.T.: Bidirectional projection network for cross dimension scene understanding. In: Proceedings of the IEEE/CVF Conference on Computer Vision and Pattern Recognition. pp. 14373--14382 (2021)

\bibitem{liteflownet}
Hui, T.W., Tang, X., Loy, C.C.: Liteflownet: A lightweight convolutional neural network for optical flow estimation. In: Proceedings of the IEEE conference on computer vision and pattern recognition. pp. 8981--8989 (2018)

\bibitem{videochat}
Li, K., He, Y., Wang, Y., Li, Y., Wang, W., Luo, P., Wang, Y., Wang, L., Qiao, Y.: Videochat: Chat-centric video understanding. arXiv preprint arXiv:2305.06355  (2023)

\bibitem{rouge}
Lin, C.Y.: Rouge: A package for automatic evaluation of summaries. In: Text summarization branches out. pp. 74--81 (2004)

\bibitem{liu2023improved}
Liu, H., Li, C., Li, Y., Lee, Y.J.: Improved baselines with visual instruction tuning. In: NeurIPS 2023 Workshop on Instruction Tuning and Instruction Following (2023)

\bibitem{videoretieval1}
Luo, H., Ji, L., Zhong, M., Chen, Y., Lei, W., Duan, N., Li, T.: Clip4clip: An empirical study of clip for end to end video clip retrieval and captioning. Neurocomputing  \textbf{508},  293--304 (2022)

\bibitem{videochatgpt}
Maaz, M., Rasheed, H., Khan, S., Khan, F.S.: Video-chatgpt: Towards detailed video understanding via large vision and language models. arXiv preprint arXiv:2306.05424  (2023)

\bibitem{PGD}
Madry, A., Makelov, A., Schmidt, L., Tsipras, D., Vladu, A.: Towards deep learning models resistant to adversarial attacks. arXiv preprint arXiv:1706.06083  (2017)

\bibitem{bleu}
Papineni, K., Roukos, S., Ward, T., Zhu, W.J.: Bleu: a method for automatic evaluation of machine translation. In: Proceedings of the 40th annual meeting of the Association for Computational Linguistics. pp. 311--318 (2002)

\bibitem{qi2023visual}
Qi, X., Huang, K., Panda, A., Wang, M., Mittal, P.: Visual adversarial examples jailbreak large language models. arXiv preprint arXiv:2306.13213  (2023)

\bibitem{radford2021learning}
Radford, A., Kim, J.W., Hallacy, C., Ramesh, A., Goh, G., Agarwal, S., Sastry, G., Askell, A., Mishkin, P., Clark, J., et~al.: Learning transferable visual models from natural language supervision. In: International conference on machine learning. pp. 8748--8763. PMLR (2021)

\bibitem{clip}
Radford, A., Kim, J.W., Hallacy, C., Ramesh, A., Goh, G., Agarwal, S., Sastry, G., Askell, A., Mishkin, P., Clark, J., et~al.: Learning transferable visual models from natural language supervision. In: International conference on machine learning. pp. 8748--8763. PMLR (2021)

\bibitem{rombach2022high}
Rombach, R., Blattmann, A., Lorenz, D., Esser, P., Ommer, B.: High-resolution image synthesis with latent diffusion models. In: Proceedings of the IEEE/CVF conference on computer vision and pattern recognition. pp. 10684--10695 (2022)

\bibitem{videocaption4}
Seo, P.H., Nagrani, A., Arnab, A., Schmid, C.: End-to-end generative pretraining for multimodal video captioning. In: Proceedings of the IEEE/CVF Conference on Computer Vision and Pattern Recognition. pp. 17959--17968 (2022)

\bibitem{LLaMA}
Touvron, H., Lavril, T., Izacard, G., Martinet, X., Lachaux, M.A., Lacroix, T., Rozi{\`e}re, B., Goyal, N., Hambro, E., Azhar, F., et~al.: Llama: Open and efficient foundation language models. arXiv preprint arXiv:2302.13971  (2023)

\bibitem{cider}
Vedantam, R., Lawrence~Zitnick, C., Parikh, D.: Cider: Consensus-based image description evaluation. In: Proceedings of the IEEE conference on computer vision and pattern recognition. pp. 4566--4575 (2015)

\bibitem{videocaption1}
Venugopalan, S., Rohrbach, M., Donahue, J., Mooney, R., Darrell, T., Saenko, K.: Sequence to sequence-video to text. In: Proceedings of the IEEE international conference on computer vision. pp. 4534--4542 (2015)

\bibitem{wang2024visionllm}
Wang, W., Chen, Z., Chen, X., Wu, J., Zhu, X., Zeng, G., Luo, P., Lu, T., Zhou, J., Qiao, Y., et~al.: Visionllm: Large language model is also an open-ended decoder for vision-centric tasks. Advances in Neural Information Processing Systems  \textbf{36} (2024)

\bibitem{sparse}
Wei, X., Zhu, J., Yuan, S., Su, H.: Sparse adversarial perturbations for videos. In: Proceedings of the AAAI Conference on Artificial Intelligence. vol.~33, pp. 8973--8980 (2019)

\bibitem{sceneunderstanding3}
Wu, Y.H., Liu, Y., Zhan, X., Cheng, M.M.: P2t: Pyramid pooling transformer for scene understanding. IEEE Transactions on Pattern Analysis and Machine Intelligence  (2022)

\bibitem{MSVDQA}
Xu, D., Zhao, Z., Xiao, J., Wu, F., Zhang, H., He, X., Zhuang, Y.: Video question answering via gradually refined attention over appearance and motion. In: Proceedings of the 25th ACM international conference on Multimedia. pp. 1645--1653 (2017)

\bibitem{xue2022advancing}
Xue, H., Hang, T., Zeng, Y., Sun, Y., Liu, B., Yang, H., Fu, J., Guo, B.: Advancing high-resolution video-language representation with large-scale video transcriptions. In: Proceedings of the IEEE/CVF Conference on Computer Vision and Pattern Recognition. pp. 5036--5045 (2022)

\bibitem{yang2024cheating}
Yang, D., Bai, Y., Jia, X., Liu, Y., Cao, X., Yu, W.: Cheating suffix: Targeted attack to text-to-image diffusion models with multi-modal priors. arXiv preprint arXiv:2402.01369  (2024)

\bibitem{videollama}
Zhang, H., Li, X., Bing, L.: Video-llama: An instruction-tuned audio-visual language model for video understanding. arXiv preprint arXiv:2306.02858  (2023)

\bibitem{videocaption3}
Zhang, Z., Qi, Z., Yuan, C., Shan, Y., Li, B., Deng, Y., Hu, W.: Open-book video captioning with retrieve-copy-generate network. In: Proceedings of the IEEE/CVF conference on computer vision and pattern recognition. pp. 9837--9846 (2021)

\bibitem{zhao2024evaluating}
Zhao, Y., Pang, T., Du, C., Yang, X., Li, C., Cheung, N.M.M., Lin, M.: On evaluating adversarial robustness of large vision-language models. Advances in Neural Information Processing Systems  \textbf{36} (2024)

\bibitem{zhu2023minigpt}
Zhu, D., Chen, J., Shen, X., Li, X., Elhoseiny, M.: Minigpt-4: Enhancing vision-language understanding with advanced large language models. arXiv preprint arXiv:2304.10592  (2023)

\bibitem{zhuang2023pilot}
Zhuang, H., Zhang, Y., Liu, S.: A pilot study of query-free adversarial attack against stable diffusion. In: Proceedings of the IEEE/CVF Conference on Computer Vision and Pattern Recognition. pp. 2384--2391 (2023)

\bibitem{zong2024safety}
Zong, Y., Bohdal, O., Yu, T., Yang, Y., Hospedales, T.: Safety fine-tuning at (almost) no cost: A baseline for vision large language models. arXiv preprint arXiv:2402.02207  (2024)

\end{thebibliography}

\clearpage 

\appendix
  In this Appendix, we provide further implementation details in Section \ref{sec:Detail}, including flow-based temporal mask, threat model specifics, datasets, and experimental setups. Following that, we present additional experimental results in Section \ref{sec:AddEXP}, featuring extended video question-answer methods (VideoChat~\cite{videochat}) and comprehensive evaluation metrics. We also report our findings from transfer-based black-box attack experiments in Section \ref{sec:Black}, showcasing the impressive transferability of our attack method. Moreover, we offer more visual comparisons in Section \ref{sec:AddVIS}. Ethics statement and reproducibility statement can be found in Section~\ref{sec:Ethics} and Section~\ref{sec:Reproducibility}, respectively. Finally, we discuss the limitations of this paper in Section~\ref{sec:Limitation}.

\section{Implementation Details}
    \label{sec:Detail}
    \noindent \textbf{Models.} We assess open-source and state-of-the-art video-based LLMs such as Video-ChatGPT~\cite{videochatgpt} and VideoChat~\cite{videochat}, ensuring reproducibility of our results. Video-ChatGPT is a multi-modal model that seamlessly combines a video-adapted visual encoder (CLIP~\cite{clip}) with a LLM, which is proficient in comprehending and generating intricate conversations related to videos. 
    VideoChat integrates video foundation models and large language models via a learnable neural interface.

    \begin{figure}[ht]
        \centering
        \includegraphics[width=0.9\textwidth]{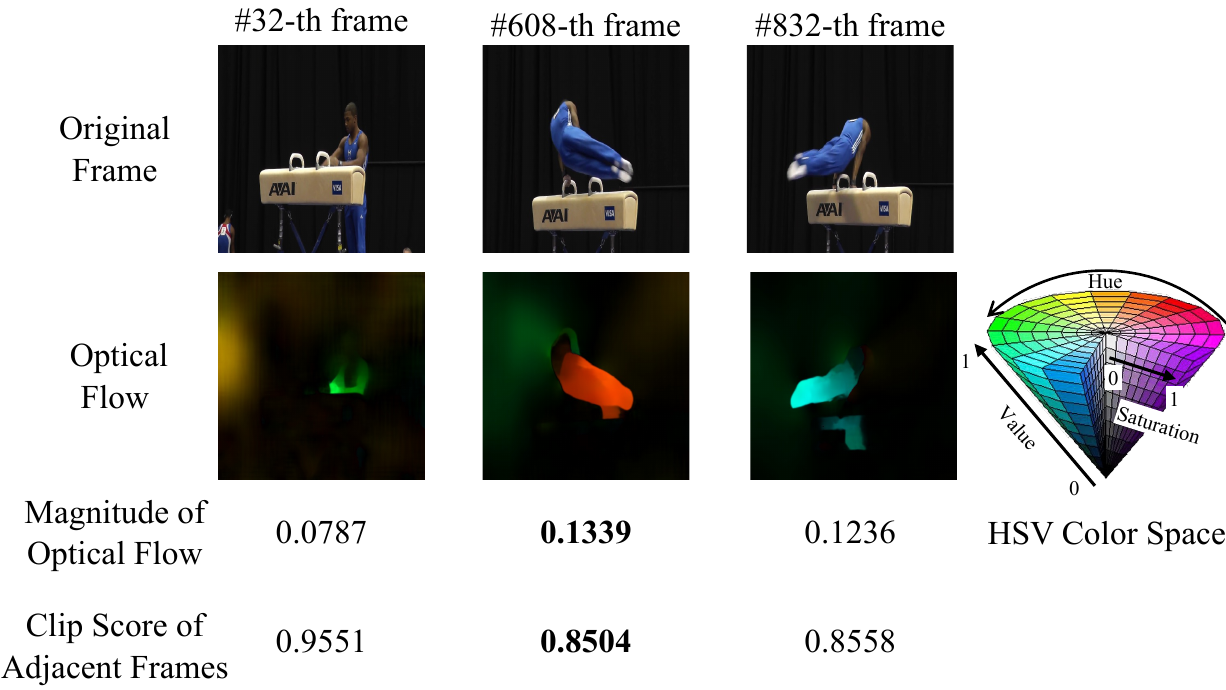}
        \caption{Relationship between optical flow and key frames. `Clip Score of Adjacent Frames' describes the similarity between the current frame and its adjacent frames, the smaller this score is the more different the current frame is. \textbf{The frames selected by flow-based masks in our FMM-Attack are key frames in the video.}}
        \label{fig:flow_inter_ori}
    \end{figure}
    
    \vspace{0.3cm}
    \noindent \textbf{Flow-based Temporal Mask.} In addition to the statistical analyses presented in the main manuscript, we visualize flow-based methods to enhance interpretability. As depicted in Fig.~\ref{fig:flow_inter_ori}, the brightness represents the magnitude of the optical flow, and the color indicates the motion direction. The motion flow magnitude varies across different frames, with a larger optical flow signifying more significant motion. Our FMM method tends to select the video frames with the largest optical flow as the key frame of the video. In other words, we tend to prioritize frames with substantial motion changes. Furthermore, the frames we select exhibit low similarity with their neighboring frames, indicating their importance.
    
    Algorithm~\ref{alg:top_k_frames} delineates the process of generating the selected set $U$ from the total set $S$. The optical flow is computed using a pre-trained liteflownet~\cite{liteflownet}.
    For the random temporal mask, we generate the selected set $U$ by randomly selecting $K$ elements from the total set $S$. Here, $S$ represents the set of frame indices, $S = \{1, 2, \ldots, T\}$, where $T$ denotes the total number of frames in the video.
    For the sequence temporal mask, we construct the selected set $U$ by sequentially selecting $K$ elements from the total set $S$. In this scenario, $U$ comprises a sequence of frames from total frames $S$, such as $\{1, 2, \ldots, K\}$ or $\{T-K+1, T-K+2, \ldots, T\}$, depending on the selected starting point within $S$.
        
    \begin{algorithm}
        \caption{Select Top K Frames with Maximum Flow}
        \label{alg:top_k_frames}
        \begin{algorithmic}[1]
        \REQUIRE video\_frames $\in T \times C \times H \times W$, $S = \{1, 2, \ldots, T\}$
        \ENSURE U, a subset with $K$ elements within $S$
        \FOR{each frame in video\_frames}
            \STATE Compute optical flow between adjacent frames
        \ENDFOR
        \FOR{each optical flow}
            \STATE Convert optical flow to color and magnitude components
            \STATE Normalize the magnitude component
        \ENDFOR
        \STATE Sort the frames based on the average value of the magnitude component
        \STATE Select the top K frame indices with the highest flow values as the set U
        \end{algorithmic}
    \end{algorithm}

    \noindent \textbf{Datasets.} In line with the Video-ChatGPT~\cite{videochatgpt} methodology, we curate a test set based on the ActivityNet-200~\cite{activitynet} and MSVD-QA~\cite{MSVDQA} datasets, featuring videos with rich, detailed descriptive captions and associated question-answer pairs obtained from human annotations. Utilizing this test set to generate adversarial examples, we effectively and quantitatively assess the adversarial robustness of video-based LLMs.

    \noindent \textbf{Experimental setups.} For evaluation, we design three spatial baselines, including videos with random perturbations, black videos with all pixel values set to 0, and white videos with all pixel values set to 1. In addition, we compare our proposed flow-based temporal mask with two straightforward temporal mask methods, serving as temporal mask baselines: the sequence temporal mask and the random temporal mask. Specifically, the sequence temporal mask consists of a continuous sequence of frame indices, while the random temporal mask comprises a randomly chosen sequence of frame indices. We utilize a variety of evaluation metrics to assess the robustness of the models.
    CLIP~\cite{clip} score characterizes the semantic similarity between the adversarial answer and the ground-truth answer. A lower CLIP score signifies a lower semantic correlation between the adversarial answer and the ground-truth answer, indicating a more effective attack. Various Image Captioning metrics such as BLEU~\cite{bleu}, ROUGE-L~\cite{rouge}, and CIDEr~\cite{cider} are used to evaluate the quality of the adversarial answer generated by the model. A lower score corresponds to a more effective attack.
    BLEU measures the overlap of n-grams between the generated and reference captions. ROUGE-L computes the longest common subsequence between them, reflecting their sentence-level similarity. CIDEr emphasizes the importance of semantically meaningful words in the captions.
    Following Video-ChatGPT~\cite{videochatgpt} and Video-LLaMA~\cite{videollama}. We also employ an evaluation pipeline using the GPT-3.5 and GPT-4 models. The pipeline employs GPT to assign a score from 1 to 5, evaluating the similarity between the output sentence and the ground truth, and a binary score (0 or 1) to measure its accuracy.

 \begin{table}[t]
            \small
            \centering
            \caption{White-box attacks against VideoChat~\cite{videochat} on the ActivityNet-200~\cite{activitynet} dataset and the MSVD-QA~\cite{MSVDQA} dataset: Comparison of image caption metrics and GPT score for different attack types. Random spatial attack denotes random perturbations added to video frames, and Black spatial attack and White spatial attack denote video frames being all 0 and all 1, respectively. The sparsity of the temporal mask is set to 0. ${\Delta}$: the mean of the modified pixels.}
            
            \begin{tabular}{c|c|c|c|c|c|c|c|c|c}
            \hline 
            \multirow{2}{*}{Dataset} & \multirow{2}{*}{Type} & \multirow{2}{*}{${\Delta}$} & \multicolumn{3}{|c}{Image Caption $\downarrow$} & \multicolumn{2}{|c|}{GPT-3.5 $\downarrow$} & \multicolumn{2}{|c}{GPT-4 $\downarrow$} \\
            \cline{4-10} 
            & & & BLEU & ROUGE & CIDEr & Accurate & Score & Accurate & Score \\
            \hline
            \multirow{5}{*}{ActivityNet~\cite{activitynet}}
            &Clean  & 0   & 0.0765 & 0.3358 & 0.3379 & 0.35 & 2.87 & 0.28 & 1.79\\
            &Random & 8   & 0.0870 & 0.3446 & 0.4225 & 0.47 & 3.04 & 0.34 & 2.06\\
            &Black  & 110 & 0.0726 & 0.3104 & 0.3599 & 0.16 & 2.17 & 0.10 & 0.66\\
            &White  & 152 & 0.0828 & 0.3226 & 0.3698 & 0.18 & 2.25 & 0.11 & 0.79\\
            &\textbf{FMM}   & 2    & \textbf{0.0626} & \textbf{0.2474} & \textbf{0.2859} & \textbf{0.06} & \textbf{1.43} & \textbf{0.06} & \textbf{0.40}\\
            \hline
            \multirow{5}{*}{MSVD-QA~\cite{MSVDQA}}
            &Clean  & 0   & 0.0415 & 0.2643 & \textbf{0.1183} & 0.70 & 3.66 & 0.68 &3.20\\
            &Random & 8   & 0.0594 & 0.2679 & 0.2182 & 0.64 & 3.44 & 0.54 &2.78\\
            &Black  & 110 & 0.0479 & 0.2674 & 0.2185 & 0.38 & 2.80 & 0.26 &1.60\\
            &White  & 153 & 0.0594 & 0.2734 & 0.4154 & 0.44 & 3.00 & 0.20 &1.46\\
            &\textbf{FMM}   & 2   & \textbf{0.0388} & \textbf{0.2001} & {0.2554} & \textbf{0.12} & \textbf{1.58} & \textbf{0.12} & \textbf{1.42}\\
            \hline
            \end{tabular}
            \label{tab:VideoChatMain}
        \end{table}

\section{Additional Experimental Results}
    \label{sec:AddEXP}
    In this section, we present additional experiments. Firstly, we execute an attack on another video-based LLM model, VideoChat~\cite{videochat}. Following that, we provide experimental results from other datasets and introduce more comprehensive evaluation metrics. Finally, we investigate the impact of varying the weights of the features loss.

    \vspace{0.3cm}
    \noindent \textbf{Additional Video-based LLMs.} In addition to the attack on Video-ChatGPT~\cite{videochatgpt} described in the main text, we also attack VideoChat~\cite{videochat}, as illustrated in Table~\ref{tab:VideoChatMain}. Our proposed FMM significantly diminishes VideoChat's question-answering capability, resulting in gibberish outputs. This is substantiated by a notable decrease in both image caption scores and GPT scores. Specific visualization results are provided in subsequent sections.

        \begin{table}[t]
            \small
            \centering
            \caption{White-box attacks against surrogate model Video-ChatGPT~\cite{videochatgpt} on the MSVD-QA~\cite{MSVDQA} dataset: Comparison of CLIP Score and Image Captioning Metrics for different attack types. Random attack denotes random perturbations added to video frames, Black attack and White attack denote video frames being all 0 and all 1, respectively. seq: sequence temporal mask. random: random temporal mask. flow: flow-based temporal mask. $M_{spa}$: Sparsity of temporal mask. ${\Delta}$: the mean of the modified pixels. }
            \begin{tabular}{c|c|c|c|c|c|c|c|c|c}
            \hline 
            \multirow{2}{*}{Type} & \multirow{2}{*}{${\Delta}$} & \multicolumn{5}{|c|}{Clip Score $\downarrow$} & \multicolumn{3}{|c}{Image Caption $\downarrow$} \\
            \cline{3-10} 
            & & RN50 & RN101 & ViT-B/16 & ViT-B/32 & ViT-L/14 & BLEU & ROUGE & CIDEr\\
            \hline
            Clean  & 0   & 0.8322 & 0.8180 & 0.8299 & 0.8533 & 0.8675 & 0.3864 & 0.6843 & 3.6792\\
            Random & 8   & 0.8249 & 0.8141 & 0.8299 & 0.8376 & 0.7446 & 0.4107 & 0.7042 & 4.0715\\
            Black  & 100 & 0.8145 & 0.7902 & 0.8334 & 0.8376 & 0.7280 & 0.3548 & 0.6478 & 3.4367\\
            White  & 148 & 0.8057 & 0.8090 & 0.8390 & 0.8435 & 0.7580 & 0.3969 & 0.6736 & 3.9245\\
            \textbf{FMM}   & 8   & \textbf{0.7337} & \textbf{0.7181} & \textbf{0.7645} & \textbf{0.7796} & \textbf{0.6460} & \textbf{0.3240} & \textbf{0.5746} & \textbf{3.1420}\\
            \hline 
            \end{tabular}
            \label{tab:MSVD}
        \end{table}
        
    \vspace{0.3cm}
    \noindent \textbf{Comprehensive Evaluation Metrics.} Owing to space constraints in the main text, we include more extensive experimental results in Table~\ref{tab:MSVD} and Table~\ref{tab:Time}. Table 2 contains five evaluation metrics related to Clip score and three metrics associated with image caption. These eight metrics display a consistent pattern, demonstrating that our approach significantly reduces the correlation between questions and answers. As a result, the scores experience a considerable decline, indicating the effectiveness of our proposed method in disrupting the performance of the targeted models.

    Table~\ref{tab:Time} includes a comparison of the temporal mask. We compare the four baselines - `Clean', `Random', `Black', and `White' - when the temporal mask is set to 0. As evident from the table, our method significantly outperforms the two attack methods, `Black' and `White', even though we only modify 8 pixel values compared to their modification of more than 100 pixel values. Additionally, when the temporal mask is not set to 0, we establish two alternative comparison methods: sequential (`seq') and `random'. The `seq' method involves inputting frames with consecutive masks, while the `random' method requires inputting frames with randomly assigned masks. Our proposed method, FMM, is based on the maximum flow algorithm (see Algorithm~\ref{alg:top_k_frames}). It selects frames corresponding to the top K largest flows according to their flow magnitude.
        
        \begin{table}[t]
            \scriptsize
            \centering
            \caption{White-box attacks against surrogate model Video-ChatGPT~\cite{videochatgpt} on the ActivityNet-200~\cite{activitynet} dataset: Comparison of CLIP Score and Image Captioning Metrics for different attack types. Random attack denotes random perturbations added to video frames, Black attack and White attack denote video frames being all 0 and all 1, respectively. seq: sequence temporal mask. random: random temporal mask. flow: flow-based temporal mask. ${M}_{spa}$: Sparsity of temporal mask. ${\Delta}$: the mean of the modified pixels. seq: sequence temporal mask. random: random temporal mask. flow: flow-based temporal mask. ${M}_{spa}$: Sparsity of temporal mask. ${\Delta}$: the mean of the modified pixels.}
            \begin{tabular}{c|c|c|c|c|c|c|c|c|c|c}
            \hline 
            \multirow{2}{*}{Type} &\multirow{2}{*}{${M}_{spa}$} & \multirow{2}{*}{${\Delta}$} & \multicolumn{5}{|c|}{Clip Score $\downarrow$} & \multicolumn{3}{|c}{Image Caption $\downarrow$} \\
            \cline{4-11} 
            & & & RN50 & RN101 & ViT-B/16 & ViT-B/32 & ViT-L/14 & BLEU & ROUGE & CIDEr\\
            \hline
            Clean  &$0\%$ & 0   & 0.7817 & 0.7827 & 0.8096 & 0.8115 & 0.7231 & 0.2029 & 0.4820 & 1.6364\\
            Random &$0\%$ & 8   & 0.7637 & 0.7681 & 0.7920 & 0.7974 & 0.6909 & 0.1986 & 0.4793 & 1.5552\\
            Black  &$0\%$ & 100 & 0.7661 & 0.7676 & 0.7959 & 0.8018 & 0.7105 & 0.1691 & 0.4570 & 1.3246\\
            White  &$0\%$ & 148 & 0.7564 & 0.7534 & 0.7813 & 0.7837 & 0.6836 & 0.1689 & 0.4545 & 1.3431\\
            \textbf{FMM} &$0\%$  & 8   & \textbf{0.6211} & \textbf{0.6274} & \textbf{0.6748} & \textbf{0.7036} & \textbf{0.5435} & \textbf{0.1336} & \textbf{0.3694} & \textbf{0.9864}\\
            \hline \hline
            
            seq &$20\%$ & 7 & 0.7500 & 0.7568 & 0.7856 & 0.7905 & 0.7041 & 0.1572 & 0.4073 & 1.2312\\
            random &$20\%$ & 7 & 0.7142 & 0.7251 & 0.7564 & 0.7637 & 0.6436 & 0.1590 & \textbf{0.3996} & 1.2858\\
            \textbf{FMM} &${20\%}$ & {7} & \textbf{0.6821} & \textbf{0.7085} & \textbf{0.7163} & \textbf{0.7446} & \textbf{0.6314} & \textbf{0.1527} & {0.4109} & \textbf{1.1696}\\
            \hline
            seq &$40\%$ & 5 & 0.7578 & 0.7500 & 0.7749 & 0.7954 & 0.7012 & 0.1606 & 0.4323 & 1.1940\\
            random &$40\%$ & 5 & 0.7559 & 0.7588 & 0.7847 & 0.7925 & 0.6914 & 0.1751 & 0.4458 & \textbf{1.4057}\\
             \textbf{FMM} &${40\%}$ & {5} & \textbf{0.7460} & \textbf{0.7559} & \textbf{0.7759} & \textbf{0.7896} & \textbf{0.6797} & \textbf{0.1485} & \textbf{0.4053} & \textbf{1.1327}\\
            \hline
            seq &$60\%$ & 4 & 0.7583 & 0.7764 & 0.8027 & 0.7993 & 0.7031 & 0.1900 & 0.4767 & 1.6175\\
            random &$60\%$ & 4 & 0.7671 & 0.7769 & 0.8042 & 0.8032 & 0.7158 & 0.1894 & 0.4727 & 1.4461\\
             \textbf{FMM} &${60\%}$ & {4} & \textbf{0.7510} & \textbf{0.7500} & \textbf{0.7710} & \textbf{0.7871} & \textbf{0.6714} & \textbf{0.1830} & \textbf{0.4713} & \textbf{1.4462}\\
            \hline
            seq &$80\%$ & 2 & 0.7715 & 0.7788 & 0.7964 & 0.7954 & 0.7119 & 0.2000 & 0.4779 & 1.6175\\
            random &$80\%$ & 2 & 0.7813 & 0.7827 & 0.8052 & 0.8071 & 0.7178 & 0.1957 & \textbf{0.4658} & 1.5905\\
             \textbf{FMM} &${80\%}$ & {2} & \textbf{0.7349} & \textbf{0.7637} & \textbf{0.7783} & \textbf{0.7886} & \textbf{0.6826} & \textbf{0.1940} & {0.4731} & \textbf{1.4555}\\
            \hline
            \end{tabular}
            \label{tab:Time}
        \end{table}
        
    \noindent \textbf{Hyperparameters.} We aimed to investigate the impact of varying the weights of the video features loss and LLM features loss on the attack effectiveness. As shown in Table~\ref{tab:weight}, different combinations of weights can lead to varying degrees of attack effectiveness (Note that $\lambda_1$ is fixed.). The most effective attack (lowest scores) is achieved when $\lambda_2 = 1$ and $\lambda_3 = 3$. 
        \begin{table}[t]
            \scriptsize
            \centering
            \caption{Influence of the weights on the attack effectiveness. We fix the sparsity loss weight $\lambda_1$ and vary the weights of the video features loss and LLM features loss to explore their relative relationship. Lower scores indicate better attack effectiveness.}
            \begin{tabular}{c|c|c|c|c|c|c|c|c|c}
            \hline \multirow{2}{*}{ $\lambda_2$ } & \multirow{2}{*}{ $\lambda_3$ } & \multicolumn{5}{|c|}{ Clip Score $\downarrow$} & \multicolumn{3}{|c}{ Image Captioning $\downarrow$} \\
            \cline{3-10} & & RN50 & RN101 & ViT-B/16 & ViT-B/32 & ViT-L/14 & BLEU & ROUGE-L & CIDEr \\
            \hline 1 & 1 & 0.7339 & 0.7383 & 0.7544 & 0.7656 & 0.6641 & 0.2133 & 0.4843 & 1.6687 \\
            \hline 1 & 2 & 0.7011 & 0.6665 & 0.7036 & 0.7461 & 0.6167 & 0.0602 & 0.2248 & 0.4535 \\
            \hline 1 & 3 & \textbf{0.6491} & \textbf{0.6060} & \textbf{0.6836} & \textbf{0.6968} & \textbf{0.5566} & \textbf{0.0230} & \textbf{0.1585} & \textbf{0.1601} \\
            \hline 1 & 4 & 0.7373 & 0.7192 & 0.7612 & 0.7705 & 0.6724 & 0.1710 & 0.3869 & 1.5064 \\
            \hline 1 & 5 & 0.7192 & 0.7061 & 0.7393 & 0.7539 & 0.6392 & 0.1207 & 0.3276 & 0.9869 \\
            \hline 2 & 1 & 0.6621 & 0.6499 & 0.6997 & 0.7285 & 0.5942 & 0.0624 & 0.2159 & 0.4087 \\
            \hline 3 & 1 & 0.7129 & 0.7227 & 0.7432 & 0.7534 & 0.6470 & 0.1142 & 0.3648 & 1.0183 \\
            \hline 4 & 1 & 0.7188 & 0.7217 & 0.7398 & 0.7744 & 0.6616 & 0.1670 & 0.4418 & 1.4612 \\
            \hline 5 & 1 & 0.7559 & 0.7578 & 0.7891 & 0.8018 & 0.7173 & 0.2047 & 0.4734 & 1.8045 \\
            \hline
            \end{tabular}
            \label{tab:weight}
        \end{table}
            
\section{Transfer-based Black-box Attacks}
    \label{sec:Black}
    \begin{figure}[ht]
        \centering            
        \includegraphics[width=0.95\textwidth]{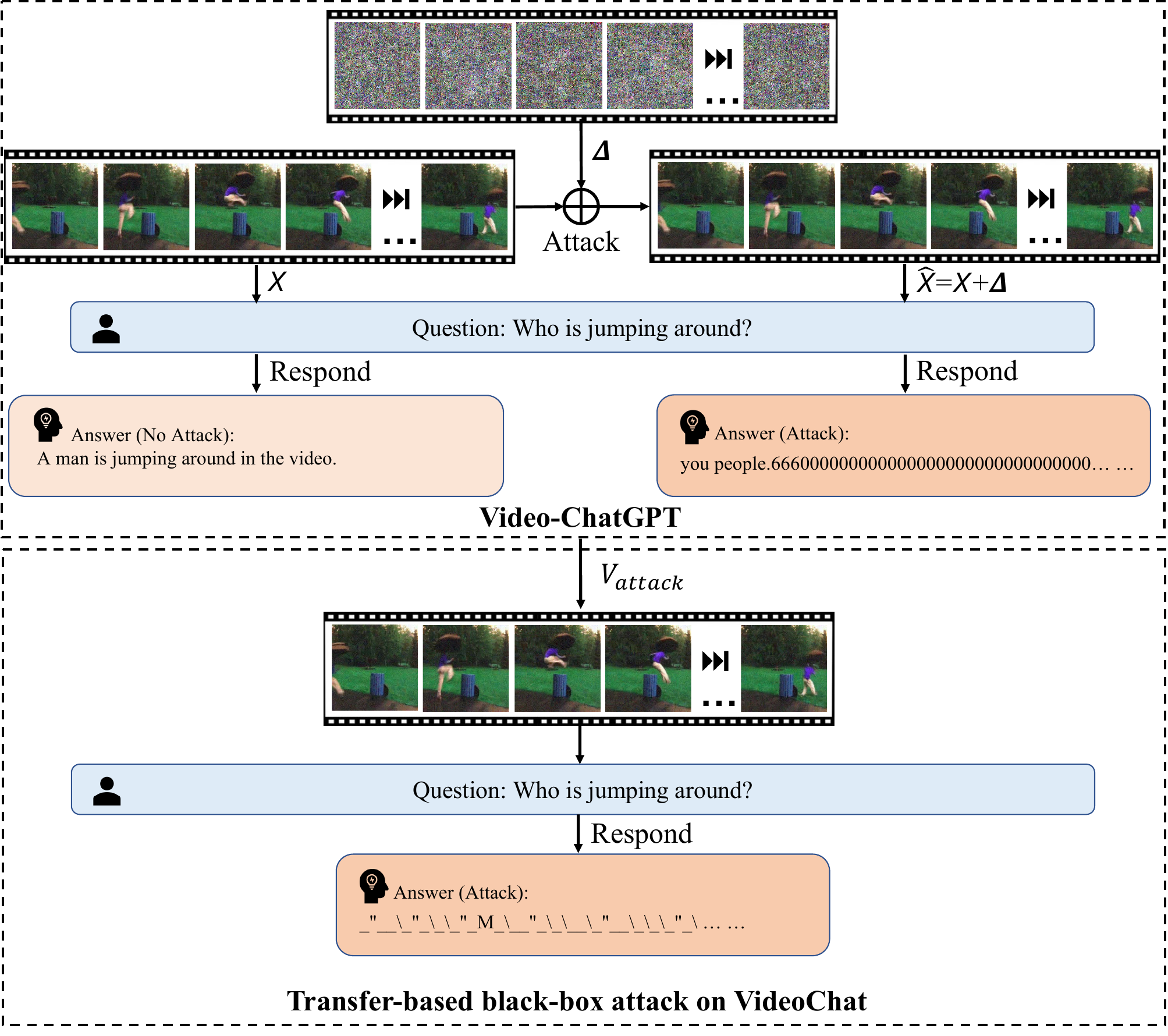}
        \caption{Transfer-based black-box attack on VideoChat.}
        \label{fig:blackbox}
    \end{figure}
    
    \begin{table}[t]
            \small
            \centering
            \caption{Black-box attacks against VideoChat~\cite{videochat} on the ActivityNet-200~\cite{activitynet} dataset: Comparison of image caption metrics and GPT score for different attack types. The sparsity of the temporal mask is set to 0. ${\Delta}$: the mean of the modified pixels. We apply the attack video on Video-ChatGPT~\cite{videochatgpt} and directly transfer it to VideoChat~\cite{videochat}.}
            
            \begin{tabular}{c|c|c|c|c|c|c|c|c}
            \hline 
            \multirow{2}{*}{Type} & \multirow{2}{*}{${\Delta}$} & \multicolumn{3}{|c}{Image Caption $\downarrow$} & \multicolumn{2}{|c|}{GPT-3.5 $\downarrow$} & \multicolumn{2}{|c}{GPT-4 $\downarrow$} \\
            \cline{3-9} 
            & & BLEU & ROUGE & CIDEr & Accurate & Score & Accurate & Score \\
            \hline
            Clean  & 0   & 0.0765 & 0.3358 & 0.3379 & 0.35 & 2.87 & 0.28 & 1.79\\
            \hline
            Transfer-based Attack  & 2    & \textbf{0.0638} & \textbf{0.2492} & \textbf{0.2870} & \textbf{0.08} & \textbf{1.56} & \textbf{0.10} & \textbf{1.32}\\
            
            \hline
            \end{tabular}
            \label{tab:Black}
        \end{table}

    In addition to white-box attacks, we have also investigated the transferability of these attacks. We conduct a black-box attack on VideoChat~\cite{videochat}. Specifically, we employ the FMM-Attack method to perform a white-box attack on Video-ChatGPT~\cite{videochatgpt}, resulting in the attack video $V_{attack}$. This video $V_{attack}$ is then directly used as input for VideoChat~\cite{videochat}, with the experimental results displayed in Table~\ref{tab:Black}. It is evident that the model's answer accuracy decreases significantly. Even without obtaining the gradient of VideoChat~\cite{videochat}, the attack is successful, and there are instances of garbled text. The visualization results is shown in Fig.~\ref{fig:blackbox}.

\section{Additional Visualization Results}
    \label{sec:AddVIS}

    In this section, we provide supplementary visualization results to further illustrate the impact of our attack method on the targeted models. The additional visualizations complement the main text's qualitative analysis, offering a more in-depth understanding of our attack's effectiveness and its implications on various models.

\begin{figure}[ht]
        \centering            
        \includegraphics[width=0.95\textwidth]{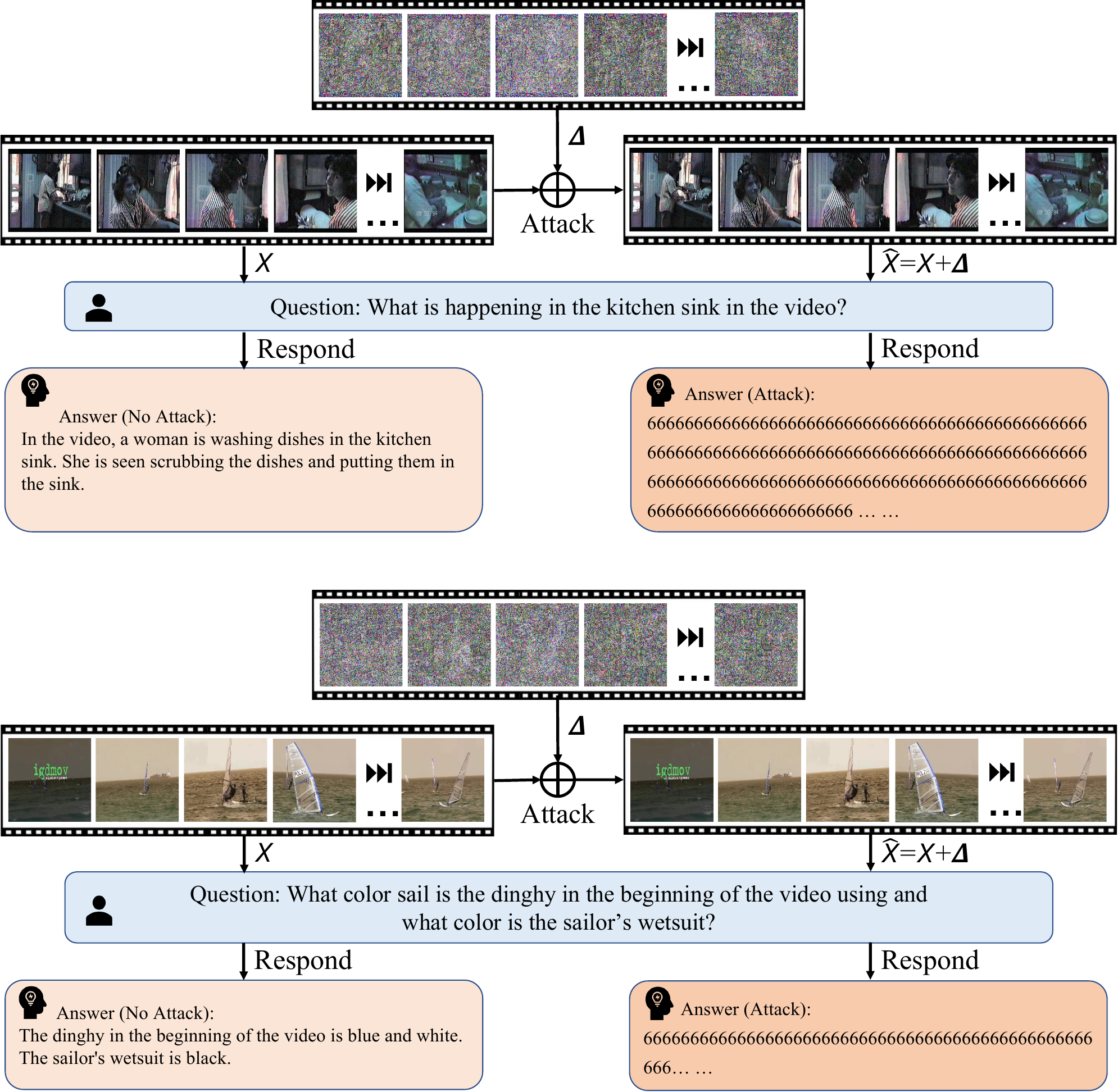}
        \caption{Perturbed videos generated by Video-ChatGPT, showcasing the impact of our attack method on the model's responses. The figure illustrates the garbled responses produced by the model, such as the repetition of the number "666666", highlighting the vulnerability of Video-ChatGPT to our adversarial attack.}
        \label{fig:vis_chatgpt}
    \end{figure}

\begin{figure}[ht]
        \centering            
        \includegraphics[width=0.95\textwidth]{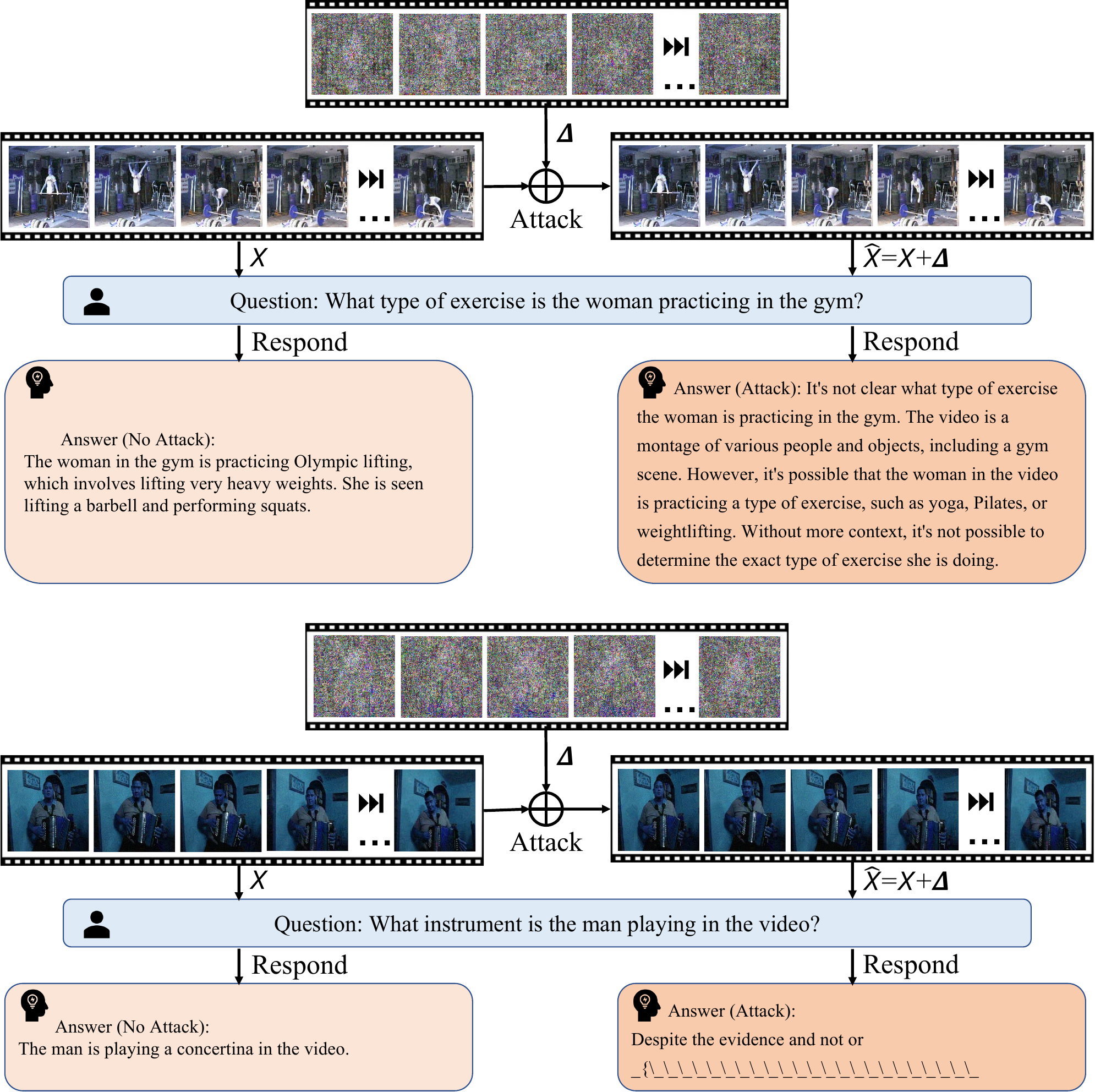}
        \caption{Perturbed videos generated by VideoChat, demonstrating the effects of our attack method on the model's responses. The figure displays the garbled responses produced by the model, including phrases like "It's not clear...", emphasizing the vulnerability of VideoChat to our adversarial attack.}
        \label{fig:vis_chat}
    \end{figure}

    As depicted in Fig.~\ref{fig:vis_chatgpt}, when attacking Video-ChatGPT\cite{videochatgpt}, the model consistently generates garbled responses, such as the repetition of the number "666666". This figure emphasizes the potency of our attack method, as it renders the model incapable of producing meaningful and contextually relevant responses. The consistent generation of garbled output highlights the vulnerability of the Video-ChatGPT model to adversarial attacks.
    In Fig.\ref{fig:vis_chat}, we present the results of our attack on VideoChat\cite{videochat}. Similar to the case of Video-ChatGPT, the model generates garbled responses, although the format differs. Notably, VideoChat sometimes tends to answer with phrases like "It's not clear...", which also signifies the success of our attack, as it effectively erases the video information. This result underlines the transferability of our attack method across different models and its ability to disrupt their performance.
    
    These additional visualization results, combined with the main text's analysis, provide a comprehensive understanding of the impact of our attack method on the targeted models. They emphasize the need for enhancing the robustness of these models against adversarial attacks and demonstrate the importance of considering different attack scenarios and their consequences. Furthermore, these results highlight the potential challenges in developing robust video question-answering systems and underscore the importance of addressing these vulnerabilities to ensure the reliability and security of such models in real-world applications.
    
\section{Ethics Statement}
\label{sec:Ethics}
Please note that we restrict all experiments in the laboratory environment and do not support our FMM-Attack in the real scenario.
The purpose of our work is to raise the awareness of the security concern in availability of video-based LLMs and call for  practitioners to pay more attention to the adversarial robustness of video-based LLMs and model trustworthy deployment.

\section{Reproducibility Statement}
\label{sec:Reproducibility}
The detailed descriptions of models, datasets, and experimental setups are provided in Section~\ref{sec:Detail}. 
We provide part of the codes to reproduce our FMM-Attack in the supplementary material. We will provide the remaining codes for reproducing our method upon the acceptance of the paper.

\section{Limitation}
\label{sec:Limitation}
Our FMM-Attack is primarily concentrated on the digital world, operating under the assumption that input videos are fed directly into the models. However, as technology advances, we anticipate that video-based LLMs will be increasingly deployed in more complex, real-world scenarios. These scenarios could include autonomous driving, where input videos are not pre-recorded but rather captured in real-time from physical environments via cameras. Future research should explore the execution and impact of adversarial attacks in the physical world. This would provide a more comprehensive evaluation of the security of video-based LLMs, contributing to the development of more robust and reliable systems for real-world deployment.

\end{document}